\documentclass[11pt, a4paper, logo, copyright]{eai}
\usepackage{shlab}

\usepackage[authoryear, sort&compress, round]{natbib}
\usepackage[]{mdframed}

\usepackage{amsmath,amsfonts,bm}
\usepackage{multirow}
\usepackage{subcaption}
\usepackage{wrapfig}

\def\eqref#1{equation~\ref{#1}}

\def\1{\bm{1}}

\DeclareMathAlphabet{\mathsfit}{\encodingdefault}{\sfdefault}{m}{sl}
\SetMathAlphabet{\mathsfit}{bold}{\encodingdefault}{\sfdefault}{bx}{n}

\usepackage{listings}
\usepackage{hyperref}
\usepackage{url}
\usepackage{graphicx}
\usepackage{amsmath} %
\usepackage{mathrsfs} %
\usepackage{etoolbox}
\usepackage{cleveref}
\usepackage{tcolorbox}
\usepackage{colortbl}
\usepackage{booktabs}       %
\usepackage{amsfonts}       %
\usepackage{nicefrac}       %
\usepackage{microtype}      %
\usepackage{caption}
\usepackage{subcaption}
\usepackage{algorithm}
\usepackage{algorithmic}
\usepackage{multirow}
\usepackage{lipsum}

\usepackage{booktabs} %
\usepackage{enumitem}%
\setlist[itemize]{noitemsep, topsep=0pt}
\usepackage{enumitem,kantlipsum}

\newlength\savewidth

\definecolor{baselinecolor}{HTML}{d6eaf8}

\definecolor{mygray}{gray}{0.4}

\AtBeginEnvironment{tcolorbox}{\tiny}

\newcount\Comments  %
\usepackage{color}
\definecolor{darkred}{rgb}{0.9,0,0}
\definecolor{darkgreen}{rgb}{0,0.5,0}
\definecolor{darkblue}{rgb}{0,0,0.7}
\definecolor{purple}{rgb}{.6, 0,.6}
\definecolor{orange}{rgb}{1.0,0.64,0}
\newcommand{\kibitz}[2]{\ifnum\Comments=1\textcolor{#1}{#2}\fi}

\usepackage{booktabs}
\usepackage{pifont}
\usepackage{xcolor}
\usepackage{wasysym}    % \smiley symbol
\usepackage{graphicx}
\usepackage{svg}
\usepackage{makecell}
\usepackage{caption}
\svgsetup{
  inkscapelatex=false
}

\newcommand{\greencheck}{\textcolor{green}{\ding{52}}} % √
\newcommand{\redcheck}{\textcolor{red}{\ding{55}}}     % ✗
\newcommand{\smileyface}{%
  \raisebox{-0.4ex}{\includegraphics[height=1.8ex]{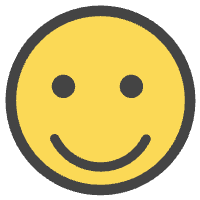}}%
}

\newcommand{\name}{{\emph{OmniWorld}}}
\newcommand{\nameGame}{{\emph{OmniWorld-Game}}}

\title{\name: A Multi-Domain and Multi-Modal Dataset for \\ 4D World Modeling}
\correspondingauthor{Corresponding author: Tong He, tonghe90@gmail.com}

\renewcommand{\today}{2025-9-16}

\makeatletter

\renewcommand\Affilfont{\fontsize{10}{12}\selectfont}
\makeatother

\author[1]{Yang Zhou}
\author[1]{Yifan Wang}
\author[1,2]{Jianjun Zhou}
\author[1]{Wenzheng Chang}
\author[1]{Haoyu Guo}
\author[1]{Zizun Li}
\author[1]{Kaijing Ma}
\author[1]{Xinyue Li}
\author[1]{Yating Wang}
\author[1]{Haoyi Zhu}
\author[1,2]{Mingyu Liu}
\author[1]{Dingning Liu}
\author[1]{Jiange Yang}
\author[1]{Zhoujie Fu}
\author[1]{Junyi Chen}
\author[2]{Chunhua Shen}
\author[1]{Jiangmiao Pang}
\author[1]{Kaipeng Zhang}
\author[1]{Tong He}

\affil[1]{Shanghai AI Lab}
\affil[2]{ZJU}
\begin{document}
\begin{abstract}
The field of 4D world modeling—aiming to jointly capture spatial geometry and temporal dynamics—has witnessed remarkable progress in recent years, driven by advances in large-scale generative models and multimodal learning.
However, the development of truly general 4D world models remains fundamentally constrained by the availability of high-quality data.
Existing datasets and benchmarks often lack the dynamic complexity, multi-domain diversity, and spatial-temporal annotations required to support key tasks such as 4D geometric reconstruction, future prediction, and camera-controlled video generation.
To address this gap, we introduce \name, a large-scale, multi-domain, multi-modal dataset specifically designed for 4D world modeling.
\name\ consists of a newly collected \nameGame\ dataset and several curated public datasets spanning diverse domains.
Compared with existing synthetic datasets, \nameGame\ provides richer modality coverage, larger scale, and more realistic dynamic interactions.
Based on this dataset, we establish a challenging benchmark that exposes the limitations of current state-of-the-art (SOTA) approaches in modeling complex 4D environments.
Moreover, fine-tuning existing SOTA methods on \name\ leads to significant performance gains across 4D reconstruction and video generation tasks, strongly validating \name\ as a powerful resource for training and evaluation.
We envision \name\ as a catalyst for accelerating the development of general-purpose 4D world models, ultimately advancing machines’ holistic understanding of the physical world.
\links{
  \link{code}{GitHub}{https://github.com/yangzhou24/OmniWorld}, 
  \link{data}{Data}{https://huggingface.co/datasets/InternRobotics/OmniWorld}, 
  \link{homepage}{Homepage}{https://yangzhou24.github.io/OmniWorld/}, 
}
% \vspace{0.5cm}
\end{abstract}

\maketitle

\begin{figure}[h!]
    \centering
    \includegraphics[width=\linewidth]{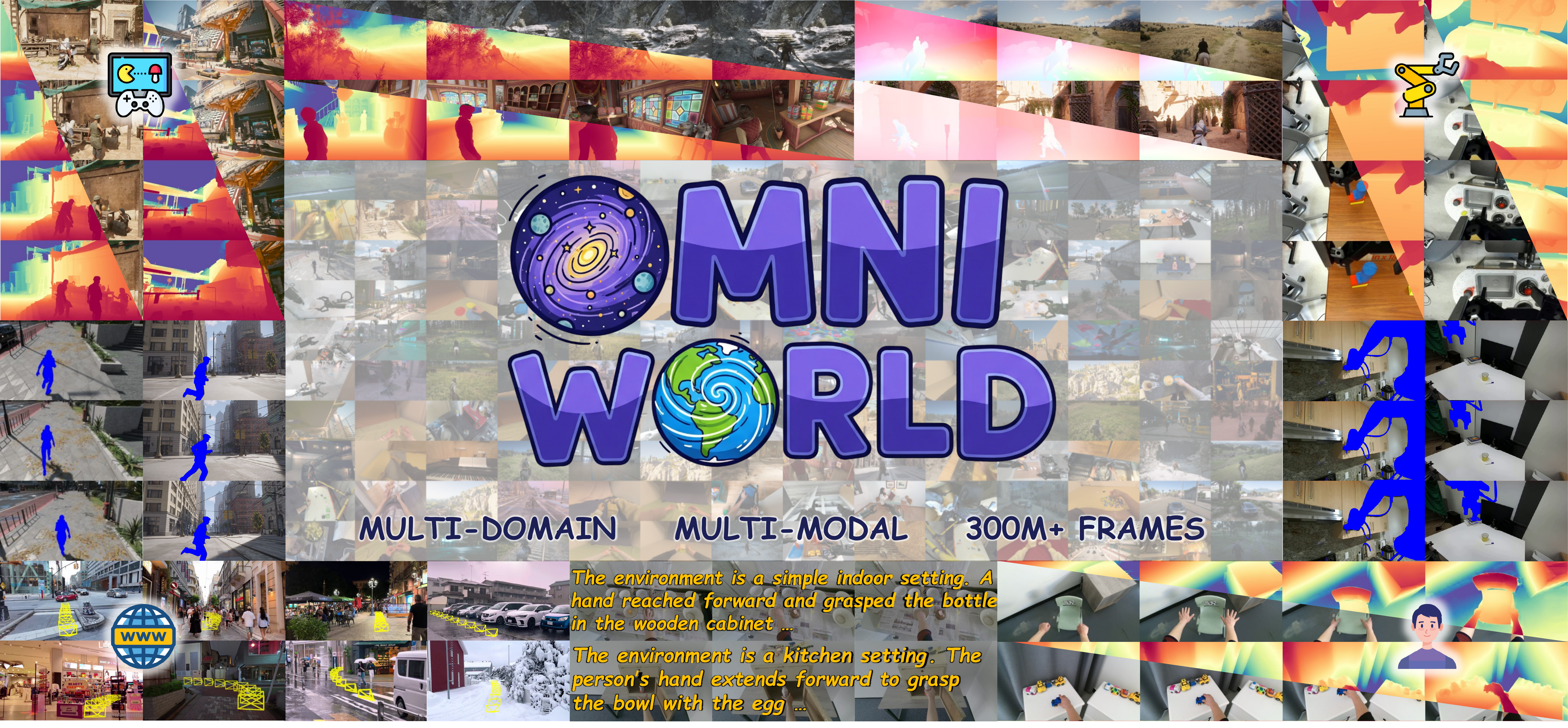}
    \captionsetup{justification=justified,singlelinecheck=false}
    \caption{
    We introduce \textbf{\name}, a large-scale, multi-domain, and multi-modal dataset. \name\ provides a rich resource for 4D world modeling by integrating high-quality data from multiple domains and offers a variety of data types, including depth maps, camera poses, text captions, optical flow and foreground masks. \name\ is designed to accelerate the development of more general models for modeling the real physical world.
    }
    \vspace{-0.5cm}
    \label{fig:first_figure_interface}
\end{figure} 

\section{Introduction}
The development of world models~\citep{genie3_2025_web, ha2018world, agarwal2025cosmos, lecun2022path, hafner2023dreamerv3} has become a central pursuit in visual intelligence systems, aiming to build systems that can simulate and reason about the physical world. This capability goes beyond simple static perception, demanding models that can simulate dynamic environments, predict object motion, infer causality, and generate content that adheres to physical laws. Such spatio-temporal modeling is a cornerstone for effective world models, with its development critically dependent on large-scale, multi-domain, and multi-modal datasets~\citep{feng2024matrixinfinitehorizonworldgeneration, aether, chen2025deepverse, he2025matrix, team2025hunyuanworld,yu2025gamefactory,yu2025context}.

Two fundamental tasks that reflect a model's world modeling capability have drawn widespread attention: 
3D geometric foundation models~\citep{dust3r_cvpr24,mast3r_eccv24,zhang2024monst3r,Yang_2025_Fast3R,tang2024mv,wang2025continuous,zhang2025flarefeedforwardgeometryappearance,wang2025vggt,wang2025pi3}, and camera-controlled video generation models~\citep{wang2024motionctrl,he2024cameractrl,zheng2024cami2v,bahmani2024ac3d,bai2025recammaster,mark2025trajectorycrafter}. The former aims to extract comprehensive 3D geometric information from 2D image inputs, while the latter focuses on generating dynamic video content that follows precise spatio-temporal instructions. Both tasks heavily rely on large-scale, high-quality datasets with rich modalities, including RGB images, depth maps, and camera poses.
\begin{table*}[t]
\captionsetup{aboveskip=2pt}
\centering
\footnotesize
\resizebox{\textwidth}{!}{%
\begin{tabular}{lcccr|ccccc}
\toprule[0.17em]
\multirow{2}{*}{Dataset} & \multirow{2}{*}{Scene Type} & \multirow{2}{*}{Motion} & \multirow{2}{*}{Resolution} & \multirow{2}{*}{\# Frames} & \multicolumn{5}{c}{Data modality}             \\
                         &                              &                         &                              &                            & Depth & Camera & Text  & Optical flow & Fg.\ masks \\ \midrule
MPI~Sintel~\citep{Butler2012ANO}               & Mixed   & Dynamic & 1024 $\times$ 436 & 1K        & \greencheck & \greencheck & \redcheck & \greencheck & \greencheck \\
FlyingThings++~\citep{Mayer_2016_CVPR,harley2022particle}      & Outdoor & Dynamic & 960  $\times$ 540 & 28K       & \greencheck & \redcheck & \redcheck  & \greencheck & \greencheck \\
TartanAir~\citep{wang2020tartanair}                & Mixed   & Dynamic & 640  $\times$ 480 & 1,000K    & \greencheck & \greencheck & \redcheck & \greencheck & \greencheck \\
BlendedMVS~\citep{yao2020blendedmvs}               & Mixed   & Static  & 768  $\times$ 576 & 17K       & \greencheck & \greencheck & \redcheck & \redcheck   & \redcheck   \\
HyperSim~\citep{roberts2021hypersim}                 & Indoor  & Static  & 1024 $\times$ 768 & 77K       & \greencheck & \greencheck & \redcheck & \redcheck   & \greencheck \\
Dynamic~Replica~\citep{karaev2023dynamicstereo}          & Indoor  & Dynamic & 1280 $\times$ 720 & 169K     & \greencheck & \greencheck & \redcheck & \greencheck & \greencheck \\
Spring~\citep{Mehl2023_Spring}                   & Mixed   & Dynamic & 1920 $\times$ 1080& 23K      & \greencheck & \greencheck & \redcheck & \greencheck & \redcheck   \\
EDEN~\citep{le21wacv}                     & Outdoor & Static  & 640  $\times$ 480 & 300K & \greencheck & \greencheck & \redcheck & \greencheck & \greencheck \\
PointOdyssey~\citep{zheng2023point}             & Mixed   & Dynamic & 960  $\times$ 540 & 216K      & \greencheck & \greencheck & \redcheck & \redcheck   & \greencheck \\
SeKai\mbox{-}Game~\citep{li2025sekai}        & Outdoor & Dynamic & 1920 $\times$ 1080& 4,320K   & \redcheck & \greencheck & \greencheck & \redcheck & \redcheck \\
\textbf{\nameGame\ (Ours)} & Mixed & Dynamic & 1280 $\times$ 720 & \textbf{18,515K}  & \greencheck & \greencheck & \greencheck & \greencheck & \greencheck \\ 
\bottomrule[0.17em]
\end{tabular}%
}% end resizebox
\captionsetup{justification=justified,singlelinecheck=false}
\caption{\textbf{Comparisons between \nameGame\ and existing synthetic datasets.} \nameGame\ surpasses existing public synthetic datasets in modal diversity and data scale.}
\label{tab:omniworld_comparison}
\end{table*}

However, existing benchmarks and datasets for evaluating and training these models have significant limitations.
In the domain of 3D geometric foundation models, existing benchmarks suffer from short sequence lengths, which constrain the evaluation of a model's long-term robustness. For example, Sintel~\citep{Butler2012ANO}, which is a widely used dataset, consists of videos with an average length of only 50 frames. Furthermore, the limited motion amplitude and single-action types within these datasets (e.g., Bonn's~\citep{palazzolo2019arxiv} focuses on indoor human motion, Kitti's~\citep{Geiger2013IJRR} focuses on outdoor street scenes) fail to comprehensively evaluate model performance in complex, dynamic environments.
Similarly, in the field of camera-controlled video generation, mainstream datasets like RealEstate10K~\citep{zhou2018stereo} primarily consist of static scenes with smooth camera trajectories. This lack of diverse object motion and complex camera operations results in a noticeable gap between the dataset's content and real-world scenarios, thereby hindering a comprehensive assessment of a model's true capabilities.

From the perspective of training data, there is a critical scarcity of high-quality, multi-domain, multi-modal datasets that include rich geometric annotations. For instance, in image or video generation, while there are numerous image-text~\citep{schuhmann2022laion5bopenlargescaledataset,gadre2023datacomp} or video-text datasets~\citep{chen2024panda70m,nan2024openvid,ju2024miradatalargescalevideodataset}, they often lack critical geometric modalities such as depth maps, camera poses, and optical flow. Similarly, the demand for large-scale, diverse datasets with accurate geometric annotations is increasingly urgent for 3D geometric foundation models. 

To address these shortcomings, we introduce \name, a large-scale, multi-domain, and multi-modal dataset composed of a self-collected high-quality \nameGame\ synthetic dataset and several public datasets. Its core characteristics are:
\textbf{1) High-Quality 4D Data.} \nameGame\ is a massive synthetic video dataset comprising over 96K clips and more than 18M frames, with a total duration of over 214 hours.
% \hy{features?}{}
It is captured from diverse game environments with 720P RGB images, dense ground truth depth maps, accurate camera poses, and annotations for text captions, optical flow and foreground masks. As shown in \Cref{tab:omniworld_comparison}, the dataset significantly surpasses existing public synthetic datasets in modal diversity and scale.
\textbf{2) Multi-Domain Coverage.} By integrating datasets from four key domains including simulator, robot, human, and the internet, \name\ covers a wide range of real-world and virtual scenarios, greatly enhancing data diversity.
\textbf{3) Multi-Modality Annotations.} \name\ provides a rich suite of multi-modal annotations, crucial for detailed world modeling, as shown in \Cref{tab:omniworld_structure}. 

\begin{table*}[t]
\captionsetup{aboveskip=2pt}
\centering
\scriptsize
\setlength{\tabcolsep}{1.8pt}
\resizebox{\textwidth}{!}{%
\begin{tabular}{lccrcc|ccccc}
\toprule[0.17em]
\multirow{2}{*}{\makecell[l]{Dataset}} &
\multirow{2}{*}{\makecell[c]{Domain}}  &
\multirow{2}{*}{\makecell[c]{\# Seq.}} &
\multirow{2}{*}{\makecell[c]{FPS}}      &
\multirow{2}{*}{\makecell[c]{Resolution}} &
\multirow{2}{*}{\makecell[c]{\# Frames}} &
\multicolumn{5}{c}{Data modality}                                         \\ 
% \cmidrule(lr){7-11}
 & & & & & & \makecell[c]{Depth} & \makecell[c]{Camera} & \makecell[c]{Text} & \makecell[c]{Opt. flow} & \makecell[c]{Fg. masks} \\ \midrule
\nameGame\                    & Simulator & 96K  & 24 & 1280×720 & 18,515K & \smileyface & \smileyface & \smileyface & \smileyface & \smileyface \\
AgiBot~\citep{bu2025agibot}                  & Robot     & 20K  & 30 & 640×480 & 39,247K  & \smileyface & \greencheck  & \greencheck  & \redcheck       & \smileyface \\
DROID~\citep{khazatsky2024droid}                   & Robot     & 35K  & 60 & 1280×720 & 26,643K & \smileyface & \greencheck  & \smileyface  & \smileyface & \smileyface \\
RH20T~\citep{fang2024rh20t}                   & Robot     &109K  & 10 & 640×360 & 53,453K  & \redcheck       & \greencheck  & \smileyface  & \smileyface & \smileyface \\
RH20T-Human~\citep{fang2024rh20t}             & Human     & 73K  & 10 & 640×360 &  8,875K  & \redcheck       & \greencheck  & \smileyface        & \redcheck       & \redcheck       \\
HOI4D~\citep{Liu_2022_CVPR}                   & Human     &  2K  & 15 & 1920×1080 &    891K &  \smileyface & \smileyface  & \smileyface  & \smileyface & \greencheck \\
Epic-Kitchens~\citep{Damen2018EPICKITCHENS}           & Human     & 15K  & 30 & 1280×720 &  3,635K  & \redcheck       & \smileyface  & \smileyface  & \redcheck       & \redcheck       \\
Ego-Exo4D~\citep{grauman2024ego}               & Human     &  4K  & 30 & 1024×1024 &  9,190K & \redcheck      & \greencheck  & \smileyface  & \smileyface       & \redcheck       \\
HoloAssist~\citep{HoloAssist2023}              & Human     &  1K  & 30 &  896×504 & 13,037K & \redcheck & \smileyface  & \smileyface  & \smileyface & \redcheck \\
Assembly101~\citep{sener2022assembly101}             & Human     &  4K  & 60 & 1920×1080 &110,831K & \redcheck       & \greencheck  & \smileyface  & \smileyface & \smileyface \\
EgoDex~\citep{hoque2025egodex}                  & Human     &242K  & 30 & 1920×1080 & 76,631K & \redcheck & \greencheck  & \smileyface  & \redcheck       & \redcheck       \\
CityWalk~\citep{li2025sekai}                & Internet  &  7K  & 30 & 1280×720 & 13,096K & \redcheck & \smileyface  & \greencheck  & \redcheck       & \redcheck       \\
\bottomrule[0.17em]
\end{tabular}%
}% end resizebox
\captionsetup{justification=justified,singlelinecheck=false}
\caption{\textbf{\name\ structure}. A smiling face (\smileyface) indicates the modality is newly (re-)annotated by us, a green check (\greencheck) denotes ground-truth data that already exists in the original dataset, and a red cross (\redcheck) marks missing modalities.}
\label{tab:omniworld_structure}
\end{table*}

Based on \nameGame, we propose a new benchmark for both 3D geometric foundation models and camera-controlled video generation models. Our \nameGame\ benchmark provides challenging, complex scenarios and dynamics that accurately reflect a model's true world capabilities, revealing the limitations of current SOTAs. By fine-tuning existing SOTAs (e.g., DUSt3R~\citep{dust3r_cvpr24}, CUT3R~\citep{wang2025continuous}, Reloc3r~\citep{reloc3r}, AC3D~\citep{bahmani2024ac3d}) with \name, we demonstrate significant performance improvements on public benchmarks. This strongly validates \name\ as a powerful training resource for enhancing world modeling capabilities.

In summary, our contributions are as follows:
\begin{enumerate}
\item We introduce \name, a multi-domain and multi-modal dataset designed to address the lack of diversity in existing datasets. Its self-collected subset, \nameGame, surpasses current synthetic datasets in both modality diversity and data volume.

\item We establish a comprehensive benchmark for 3D geometric foundation models and camera-controlled video generation models based on \nameGame, providing a unified platform for evaluation.

\item We fine-tune several SOTAs on \name\ and observe significant performance gains, underscoring its value as a training resource.
\end{enumerate}

\begin{figure*}[tb]
    \centering
    % \vspace{-1em}
    \includegraphics[width=\textwidth]{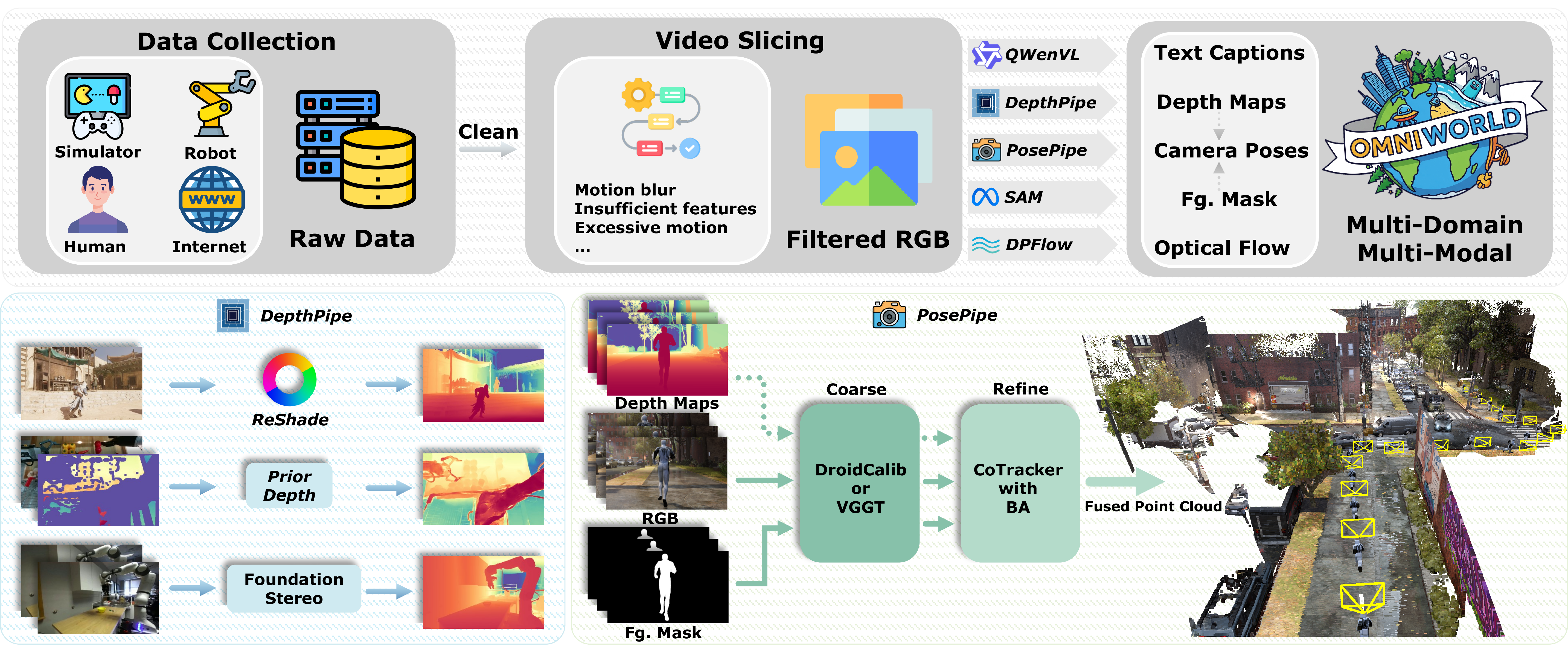}
    \captionsetup{justification=justified,singlelinecheck=false}
    \caption{~\textbf{\name\ acquisition and annotation pipeline.} We collect raw data from diverse domains and apply a video slicing filter to obtain high-quality RGB sequences. These sequences are then processed through a suite of specialized pipelines to generate multi-modal annotations, including text captions, depth maps, camera poses, foreground masks, and optical flow.}
    \label{fig:pipeline}
    % \vspace{-1em}
\end{figure*}
\section{\name\ Dataset}
To advance comprehensive spatio-temporal modeling of the real physical world, we curate \name, a large-scale, multi-domain and multi-modal dataset that mirrors the complexity of the physical world.  We design and implement a detailed data acquisition and annotation pipeline to ensure high-quality multi-modal annotations, as illustrated in \Cref{fig:pipeline}.

\subsection{Data Acquisition}
\label{sec:acquisition}
To address the scarcity of high-precision, temporally consistent, and dynamically rich data, we develop a sophisticated data acquisition pipeline. Our approach is centered on a novel self-collected dataset, \nameGame, which we supplement with data from three other domains: robot, human, and internet. This strategy allows us to integrate the strengths of diverse data sources to comprehensively capture real-world complexity.

\noindent\textbf{Simulator Domain.}
To acquire the high-precision and temporally consistent multimodal data that is hard to obtain in the real world, we collect \nameGame\ from game environments. Following prior works~\citep{richter2016playing,yang2024depthanyvideo,feng2024matrixinfinitehorizonworldgeneration,aether}, we utilize ReShade~\citep{reshade} to access depth information during the rendering process, and simultaneously capture synchronized RGB images from the screen using OBS~\citep{OBS}. This approach offers significant advantages: 
1) High-Precision Modal Data. We can precisely control the environment and acquire accurate depth data, which is often unattainable in real-world settings and is crucial for spatio-temporal modeling.
2) Rich Real-World Scene Simulation. Modern virtual environments provide highly realistic graphics and diverse simulations of real-world scenarios, encompassing complex settings from wilderness to urban areas, and from day to night.

\noindent\textbf{Robot Domain.} We integrate public datasets from robot manipulation and human-robot interaction tasks, including AgiBot~\citep{bu2025agibot}, DROID~\citep{khazatsky2024droid}, and RH20T~\citep{fang2024rh20t}.
% \hy{remove}{see caption}
These datasets provide valuable sequences of robot-environment interactions and navigation, which are essential for tasks involving robotic manipulation and physical world understanding.

\noindent\textbf{Human Domain.} We incorporate public datasets describing various human activities, including RH20T-Human~\citep{fang2024rh20t}, HOI4D~\citep{Liu_2022_CVPR}, Epic-Kitchens~\citep{Damen2018EPICKITCHENS}, Ego-Exo4D~\citep{grauman2024ego}, HoloAssist~\citep{HoloAssist2023}, Assembly101~\citep{sener2022assembly101}, and EgoDex~\citep{hoque2025egodex}. These datasets capture diverse human behaviors, ranging from daily activities to complex assembly tasks, from both egocentric and exocentric perspectives. 

\noindent\textbf{Internet Domain.} To acquire large-scale, realistic, and diverse in-the-wild scene data, we utilize the CityWalk dataset~\citep{li2025sekai}. This dataset offers rich real-world street view videos from the internet. We specifically focus on supplementary camera pose annotation for this data, providing valuable real-world information for 3D geometry and camera pose estimation tasks.

To prepare raw data for our annotation pipeline, we first perform video slicing to ensure all clips are of high quality and temporal coherence. This process has two main objectives: first, to remove frames unsuitable for geometric or motion analysis, such as those with motion blur, insufficient feature points, or excessively large dynamic areas; and second, to segment long videos into shorter, manageable clips. After this preprocessing step, the filtered, high-quality video segments are then passed to our multi-modal annotation pipeline.

\subsection{Data Annotation}
To provide high-quality multi-modal annotation information, we design an innovative data processing pipeline. We primarily annotate the following key modalities: depth maps, camera poses, text captions, optical flow, and foreground masks (see \Cref{fig:pipeline} for the overall pipeline). These modalities are crucial for models to achieve comprehensive spatio-temporal modeling. Here we briefly introduce the annotation method of each modality, please refer to supplementary material for more details.

\noindent\textbf{Depth maps.}
Accurate depth information is paramount for geometric modeling. 
To ensure the quality and consistency of depth maps, we adopt a tailored approach based on the data source.
For the self-collected dataset \nameGame, as mentioned in \Cref{sec:acquisition}, we directly access depth information during the rendering process using tools like ReShade~\citep{reshade}. 
For public datasets AgiBot~\citep{bu2025agibot} and HOI4D~\citep{Liu_2022_CVPR}, these datasets typically provide raw depth maps that are often noisy and sparse. We employ Prior Depth Anything~\citep{wang2025depthprior} to optimize these noisy depth maps, generating denser and more accurate depth maps.
For the public stereo dataset DROID~\citep{khazatsky2024droid}, we leverage FoundationStereo~\citep{wen2025stereo} for stereo depth estimation on this dataset. 

\noindent\textbf{Foreground masks.}
To provide precise, temporally consistent masks of primary subjects for tasks like subject-environment interaction and behavior analysis, we develop specialized automated pipelines. For robot domain data, we use RoboEngine~\citep{yuan2025roboengine} to generate initial masks for keyframes, followed by temporal tracking and fusion with SAM 2~\citep{ravi2024sam2}. For \nameGame\ (e.g., player characters in third-person view), we leverage Grounding DINO~\citep{liu2023grounding} to detect initial bounding boxes within predefined regions of keyframes, which then serve as prompts for SAM~\citep{kirillov2023segany}. These generated masks can be used as dynamic foreground masks to guide camera pose estimation, as detailed in the following section.

\noindent\textbf{Camera poses.} 
Accurate camera pose annotation in dynamic videos is highly challenging due to transitions, weakly textured areas, and abrupt movements that hinder traditional Structure-from-Motion methods~\citep{rockwell2025dynamic,li2024_megasam}. Following prior work~\citep{aether}, we develop a robust, automated, two-stage pipeline for dynamic camera pose annotation, whose principles are validated across diverse data types.

% \hy{move step 1 to mask}{}
The pipeline leverages the pre-computed foreground masks to focus on static background regions.
The stages include: 1) Coarse camera pose estimation leveraging VGGT~\citep{wang2025vggt} for videos without depth or DroidCalib~\citep{hagemann2023deep} with depth constraints; 2) Camera pose refinement through dense point tracking (SIFT~\citep{lowe2004distinctive}, SuperPoint~\citep{detone2018superpoint} with CoTracker3~\citep{karaev2024cotracker3}) on static regions and subsequent bundle adjustment to minimize reprojection errors, optionally enhanced by forward-backward reprojection with depth information~\citep{chen2019self}.

\noindent\textbf{Text captions.}
We generate high-quality text descriptions for video sequences using a semi-automated approach primarily driven by Qwen2-VL-72B-Instruct model~\citep{Qwen2VL}. We design specific prompting strategies tailored to different data domains. For robot and human domain data, we first annotate overall video tasks, then annotate in units of 81-frame segments. For \nameGame, we develop distinct prompts for various viewpoints (e.g., first-person, third-person), encompassing types such as short caption, player character caption, background caption, camera caption, video caption, and key tags, utilizing 81-frame segments.

\noindent\textbf{Optical flow.}
Optical flow, as a dense motion vector field, is crucial for capturing pixel-level motion information in videos and serves as a fundamental modality for accurate spatio-temporal modeling. We select DPFlow~\citep{morimitsu2025dpflow} for optical flow annotation.
Unlike mainstream models such as RAFT~\citep{teed2020raft} which require downsampling inputs when processing high-resolution videos, DPFlow can directly perform predictions on the original resolution.
Given that our dataset includes various resolutions, the choice of DPFlow ensures that the optical flow annotation accurately reflects subtle movements within the videos.

\begin{figure*}[tb]
    \centering
    \begin{subfigure}{0.29\textwidth}
        \includegraphics[width=\linewidth]{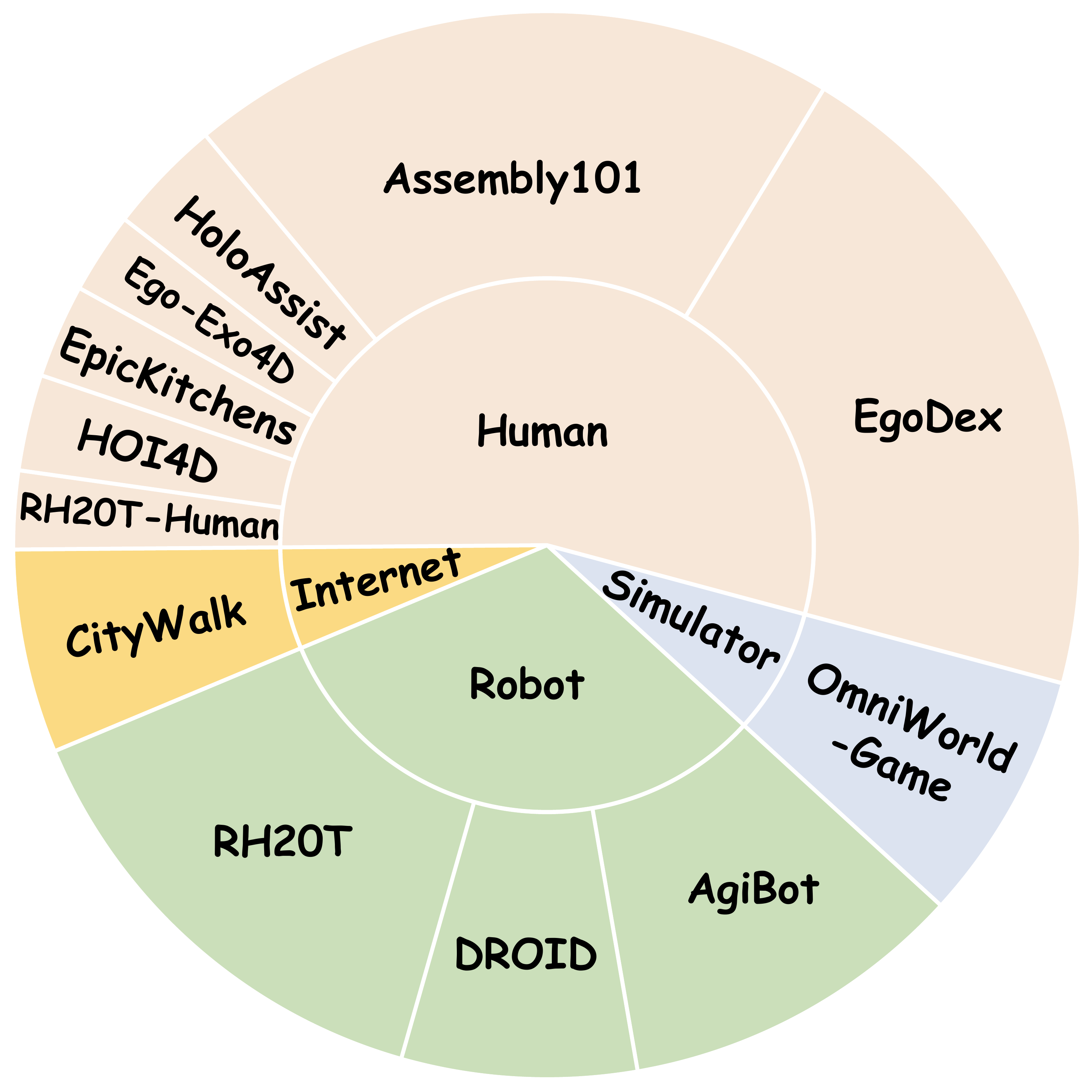}
        \vspace{-0.18 cm}
        \caption{\name\ Compositional Distribution}
        \label{fig:fig-statistics-a}
    \end{subfigure}
    \hspace{0.01\textwidth}
    \begin{subfigure}{0.28\textwidth}
        \includegraphics[width=\linewidth]{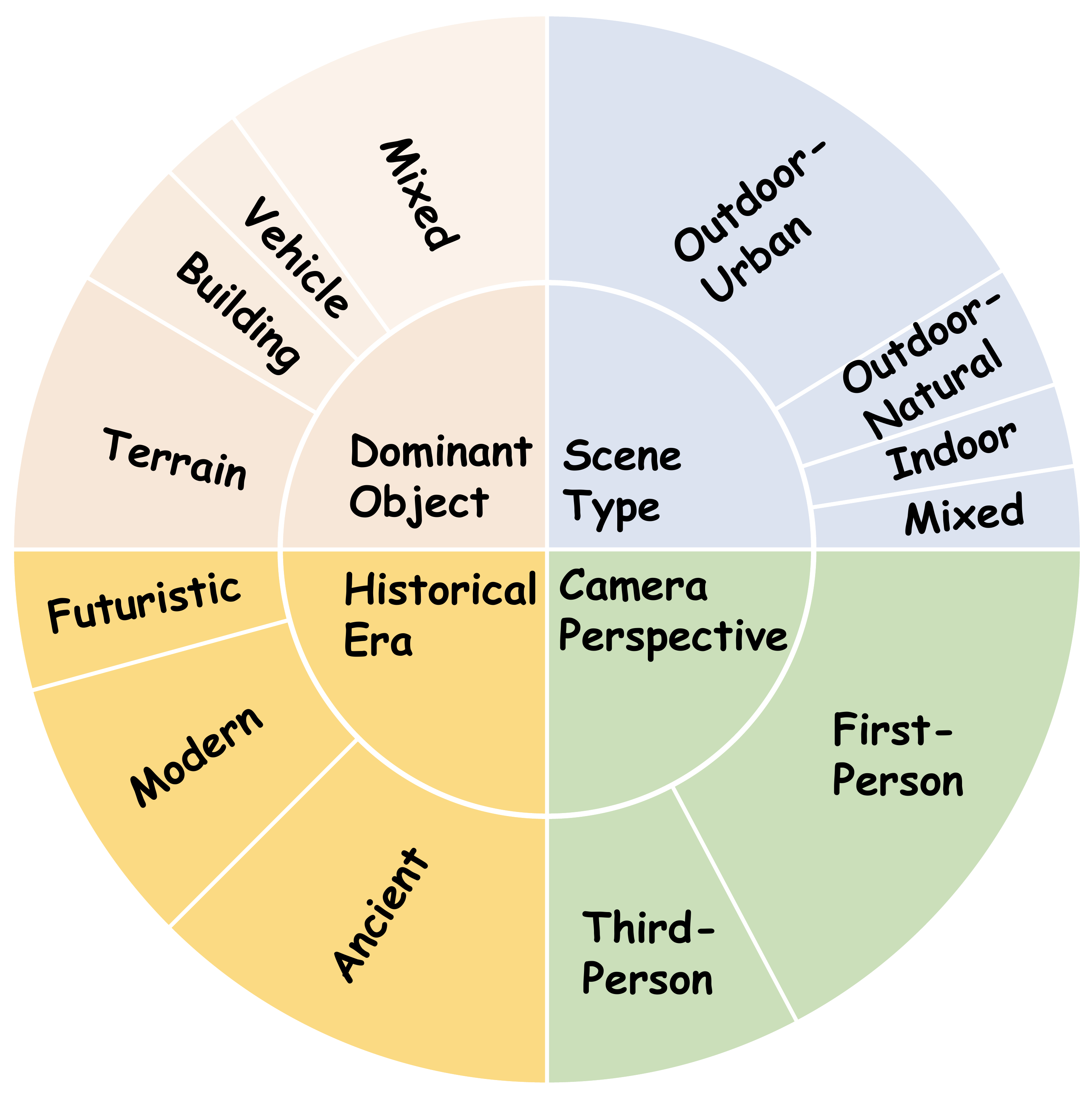}
        \vspace{-0.18 cm}
        \caption{\nameGame\ Internal Composition}
        \label{fig:fig-statistics-b}
    \end{subfigure}
    \hspace{0.01\textwidth}
    \begin{subfigure}{0.38\textwidth}
        \includegraphics[width=\linewidth]{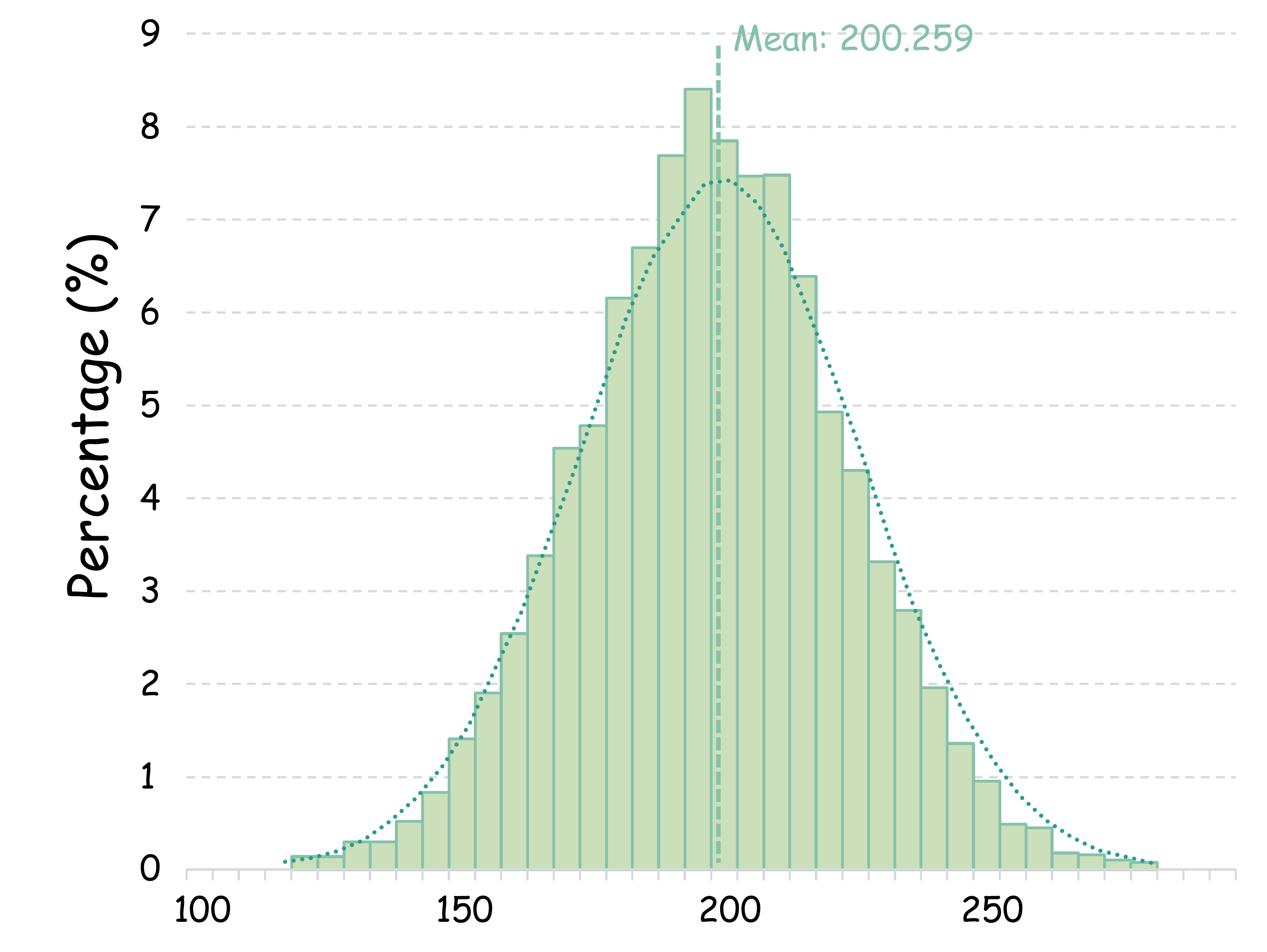}
        \vspace{-0.18 cm}
        \caption{Caption Tokens Distribution}
        \label{fig:fig-statistics-c}
    \end{subfigure}
    \captionsetup{justification=justified,singlelinecheck=false}
    \caption{\textbf{Statistical information of \name.} (a) displays compositional distribution of data from different domains within \name, (b) presents internal composition of \nameGame. (c) shows caption tokens distribution of \name.}
    \label{fig:fig-statistics}
\end{figure*}

\subsection{Data Statistics}

\name\ comprises 12 heterogeneous datasets from four domains: simulators, robots, humans, and the internet. \Cref{tab:omniworld_structure} summarizes the key metadata for these datasets.
\name\ collectively contains over 600 thousand video sequences and more than 300 million frames.  Notably, our collection includes a significant portion of high-resolution videos, with more than half of the data having a resolution of 720P or higher.
We meticulously annotate the data with multiple modalities, including depth, camera poses, text, optical flow, and foreground masks. 

\Cref{fig:fig-statistics-a} illustrates the compositional distribution of data from different domains within \name. Notably, data from the human domain constitutes the largest share, underscoring the dataset's richness in reflecting real-world human activities and interactions.

\Cref{fig:fig-statistics-b} further elaborates on the internal composition of \nameGame, showcasing its high diversity across multiple dimensions.
For scene type, \nameGame\ encompasses outdoor-urban, outdoor-natural, indoor, and mixed scenes, with outdoor-urban scenes having the highest proportion.
For camera perspective, \nameGame\ includes both first-person and third-person-following perspectives, predominantly featuring first-person views.
Regarding the historical era, \nameGame\ covers diverse styles, including ancient, modern, and futuristic sci-fi periods.
In terms of dominant object, \nameGame\ includes various types such as natural terrain, architecture, vehicles, and mixed elements. Most scenes incorporate multiple object types, significantly enhancing the data's challenge and complexity.
These statistics collectively demonstrate that \nameGame\ exhibits an exceptionally diverse and challenging scene distribution.

For the text modality, we provide comprehensive and detailed annotations. As shown in \Cref{fig:fig-statistics-c}, our text captions primarily contain between 150 and 250 tokens per description. This rich annotation density significantly surpasses that of most existing video-text datasets, such as OpenVid-1M~\citep{nan2024openvid} and Panda-70M~\citep{chen2024panda70m}.

\section{\nameGame\ Benchmark}

To comprehensively evaluate and advance world modeling, we construct \nameGame\ benchmark, providing a comprehensive and challenging evaluation platform for two critical tasks: 3D geometric prediction and camera-controlled video generation. 

\subsection{3D Geometric Prediction Benchmark}

\noindent\textbf{Benchmark design and motivation.}
Existing benchmarks for 3D geometric foundation models (GFMs) suffer from significant limitations. Specifically, many current benchmarks have the following drawbacks: First, sequence lengths are generally short, which restricts evaluating models' ability in long sequences reconstruction. For instance, Sintel~\citep{Butler2012ANO} video sequences average only 50 frames. Second, the dynamic motion in these datasets is relatively small in amplitude and uniform in type. For example, Bonn~\citep{palazzolo2019arxiv} focuses on human dynamics in indoor scenes, NYU-v2~\citep{silberman2012indoor} focuses on indoor static objects, and KITTI~\citep{Geiger2013IJRR} datasets only include outdoor street views, making it challenging to comprehensively test model performance in complex dynamic environments.

To address this, \nameGame\ offers an advanced evaluation environment featuring extended temporal sequences (up to 16 seconds with 384 frames), rich and diverse motion, extreme scenarios with environmental diversity (e.g., mixed scene types), and high-resolution realistic data (720P). These characteristics allow for a deeper and more comprehensive assessment of GFMs capabilities.

\noindent\textbf{Evaluated baselines and experiment details.}
We thoroughly assess current GFMs, including DUSt3R~\citep{dust3r_cvpr24}, MASt3R~\citep{mast3r_eccv24}, MonST3R~\citep{zhang2024monst3r}, Fast3R~\citep{Yang_2025_Fast3R}, CUT3R~\citep{wang2025continuous}, FLARE~\citep{zhang2025flarefeedforwardgeometryappearance}, VGGT~\citep{wang2025vggt}, and MoGe~\citep{wang2024moge,wang2025moge2}, within the \nameGame\ benchmark. These models are evaluated on two core tasks: monocular depth estimation and video depth estimation. All images are consistently resized to a long side of 512 pixels while preserving aspect ratio.

\noindent\textbf{Quantitative analysis.}
Our quantitative analysis on \nameGame\ reveals key performance insights and bottlenecks. For monocular depth estimation, MoGe-2 achieves the best results, though significant room for improvement remains across models, underscoring the benchmark's challenge on single-frame geometric understanding (\Cref{tab:benchmark_mono_video_depth}). In the more demanding video depth estimation task, VGGT demonstrated superior performance across all metrics under both scale-only and scale-and-shift alignments, with significantly higher FPS than competitors. While MASt3R also showed competitive metrics, its low FPS due to global alignment limits its practicality (\Cref{tab:benchmark_mono_video_depth}). Overall, no single GFM achieves top-tier performance across all metrics, indicating that current SOTAs still face considerable challenges in handling the high-dynamic, long-sequence 3D geometric understanding and consistency problems introduced by \nameGame.

\begin{figure*}[!t]
    \centering
    % \vspace{-1em}
    \captionsetup{justification=justified,singlelinecheck=false}
    \includegraphics[width=\textwidth]{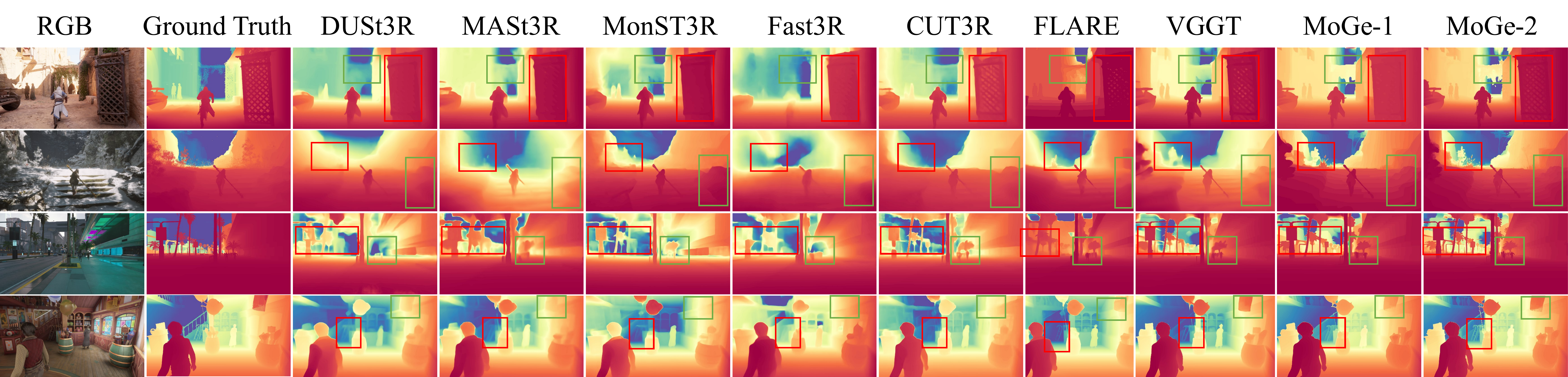}
    \caption{~\textbf{Qualitative comparison of Monocular Depth Estimation} on \nameGame\ benchmark.}
    \label{fig:benchmark_monodepth}
    % \vspace{-1em}
\end{figure*}

\noindent\textbf{Visual Results.}
In \Cref{fig:benchmark_monodepth}, we provide a visual comparison of the monocular depth prediction results from various methods on the \nameGame\ benchmark. As a model specifically designed for monocular geometry tasks, MoGe-2~\citep{wang2025moge2} achieves superior accuracy and produces visually sharp depth maps, surpassing the performance of other multi-view methods.

\begin{figure*}[!t]
    \centering
    % \vspace{-1em}
    \includegraphics[width=\textwidth]{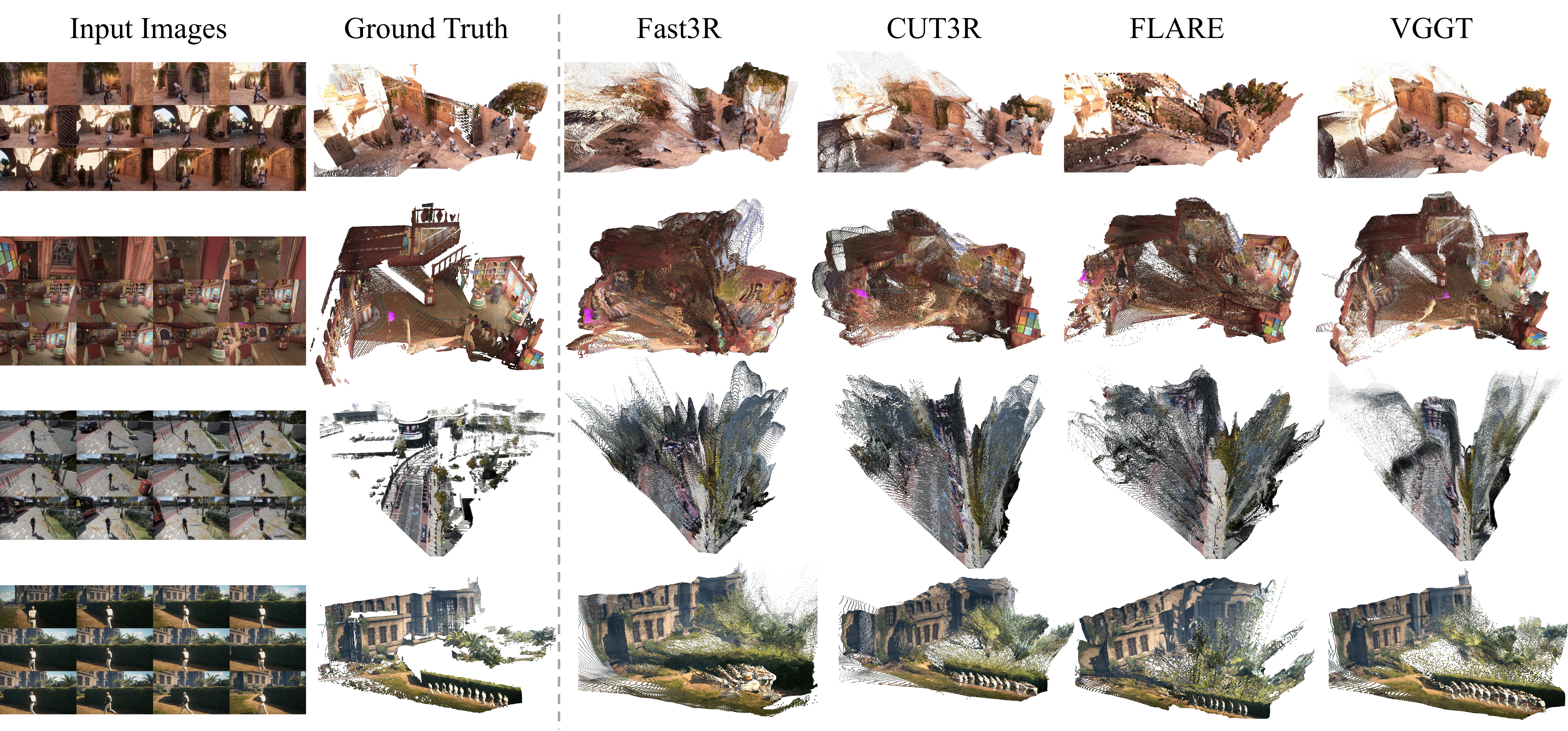}
    \captionsetup{justification=justified,singlelinecheck=false}
    \caption{~\textbf{Qualitative comparison of multi-view 3D reconstruction} on \nameGame\ benchmark. }
    \label{fig:benchmark_pointmap}
    % \vspace{-1em}
\end{figure*}

To show the challenges of video depth estimation on the \nameGame\ benchmark, we present a qualitative comparison of feed-forward reconstruction methods using point cloud visualizations in \Cref{fig:benchmark_pointmap}.
The video-depth estimation task demands high temporal consistency. Our visualizations show that VGGT~\citep{wang2025vggt} generates more coherent 3D structures than other methods in dynamic scenes. However, even VGGT shows noticeable artifacts, revealing limitations in capturing complex details.

These observations indicate that the robustness of current methods needs improvement on \nameGame. Our benchmark provides a clear direction for advancing the next generation of GFMs with stronger spatio-temporal consistency.

\begin{table}[t]
\captionsetup{aboveskip=2pt}
    \centering
    % \vspace{-1em}
    \small
    \setlength{\tabcolsep}{4.5pt}
    \resizebox{\columnwidth}{!}{%
    \begin{tabular}{lccccccc}
        \toprule[0.17em]
        % ----------------------------- first header row -------------
        \multirow{3}{*}{\centering\textbf{Method}} &
        \multicolumn{2}{c}{\textbf{Mono-Depth}} &
        \multicolumn{5}{c}{\textbf{Video-Depth}} \\
        \cmidrule(r){2-3} \cmidrule(r){4-8}
        % ----------------------------- second header row ------------
        & \multicolumn{2}{c}{scale} &
          \multicolumn{2}{c}{scale} &
          \multicolumn{2}{c}{scale\&shift} &
          \multirow{2}{*}{\centering\textbf{FPS}} \\
        \cmidrule(r){2-3} \cmidrule(r){4-5} \cmidrule(r){6-7}
        % ----------------------------- third header row -------------
        & {\footnotesize Abs Rel $\downarrow$} & {\footnotesize $\delta$\textless{}$1.25\uparrow$} &
          {\footnotesize Abs Rel $\downarrow$} & {\footnotesize $\delta$\textless{}$1.25\uparrow$} &
          {\footnotesize Abs Rel $\downarrow$} & {\footnotesize $\delta$\textless{}$1.25\uparrow$} & \\[-0.2em]
        \midrule
        % ----------------------------- data rows -------------------
        DUSt3R~\citep{dust3r_cvpr24}   & 0.742 & 0.460 & 0.709 & 0.447 & 0.379 & 0.560 & 0.96  \\
        MASt3R~\citep{mast3r_eccv24}   & 0.485 & 0.560 & \underline{0.482} & \underline{0.579} & \underline{0.217} & \underline{0.724} & 0.79  \\
        MonST3R~\citep{zhang2024monst3r}  & 0.670 & 0.493 & 0.669 & 0.505 & 0.272 & 0.648 & 0.95  \\
        Fast3R~\citep{Yang_2025_Fast3R}   & 0.755 & 0.404 & 0.741 & 0.384 & 0.464 & 0.531 & \underline{14.99} \\
        CUT3R~\citep{wang2025continuous}    & 0.624 & 0.518 & 0.690 & 0.479 & 0.429 & 0.603 & 10.75 \\
        FLARE~\citep{zhang2025flarefeedforwardgeometryappearance}    & 0.664 & 0.475 & 0.757 & 0.453 & 0.511 & 0.527 & 4.24  \\
        VGGT~\citep{wang2025vggt}     & 0.531 & 0.554 & \textbf{0.440} & \textbf{0.625} & \textbf{0.194} & \textbf{0.755} & \textbf{18.75} \\
        MoGe-1~\citep{wang2024moge}  & \underline{0.459} & \underline{0.586} &   --  &   --  &   --  &   --  &  --   \\
        MoGe-2~\citep{wang2025moge2}  & \textbf{0.401} & \textbf{0.589} &   --  &   --  &   --  &   --  &  --   \\
        \bottomrule[0.17em]
    \end{tabular}%
    }% end resizebox
    \captionsetup{justification=justified,singlelinecheck=false}
    \caption{\textbf{Monocular Depth \& Video Depth Estimation} on \nameGame\ benchmark.}
    % \vspace{-1em}
    \label{tab:benchmark_mono_video_depth}
\end{table}
\begin{table}[t]
\captionsetup{aboveskip=2pt}
\centering
\vspace{-1em}
\resizebox{1.0\columnwidth}!{
\begin{tabular}{lccccc}
\toprule[0.17em]
\multirow{2}{*}{\centering\textbf{Method}} &
\multirow{2}{*}{\centering\textbf{TransErr$\downarrow$}} &
\multirow{2}{*}{\centering\textbf{RotErr$\downarrow$}}  &
\multirow{2}{*}{\centering\textbf{CamMC$\downarrow$}}  &
\multicolumn{2}{c}{\textbf{FVD}} \\ 
\cmidrule(lr){5-6}
 &  &  &  &
 {\footnotesize VideoGPT$\downarrow$} &
 {\footnotesize StyleGAN$\downarrow$} \\
\midrule
AC3D (T2V)~\citep{bahmani2024ac3d}        & 6.2788 & 0.8867 & 6.6965 & 1745.778 & 1594.885 \\
\midrule
MotionCtrl (I2V)~\citep{wang2024motionctrl}  & 7.8633 & 1.1402 & 8.2710 & \underline{694.342} & \phantom{0}745.652 \\
CamCtrl (I2V)~\citep{he2024cameractrl}     & \textbf{1.2882} & \textbf{0.2022} & \textbf{1.3856} & \textbf{615.417} & \textbf{637.574} \\
CAMI2V (I2V)~\citep{zheng2024cami2v}      & \underline{5.9626} & \underline{0.5087 }& \underline{6.2010} & 837.185 & \underline{742.594} \\
\bottomrule[0.17em]
\end{tabular}}
\captionsetup{justification=justified,singlelinecheck=false}
\caption{\small \textbf{Camera-Controlled Video Generation Evaluation} on \nameGame\ benchmark.}
\label{tab:benchmark_camctrl_eval}
% \vspace{-1em}
\end{table}

\subsection{Camera-Controlled Video Generation Benchmark}

\noindent\textbf{Benchmark design and motivation.}
Existing benchmarks for camera-controlled video generation often rely on static datasets with smooth camera trajectories (e.g., RealEstate10K~\citep{zhou2018stereo}), which do not reflect real-world complexity. \nameGame\ benchmark addresses this by providing a challenging testing environment with rich dynamic content (e.g., diverse motions, complex interactions), extremely diverse scenes and environments (e.g., varied geographical, weather, lighting conditions), complex camera trajectories reflecting real patterns, and multi-modal input with diverse subjects (e.g., various perspectives, characters, vehicles). This enables a rigorous evaluation of models' ability to handle complex spatio-temporal dynamics and adhere to precise control instructions. 

\noindent\textbf{Evaluated baselines and experiment details.}
We benchmark mainstream SOTAs, including AC3D~\citep{bahmani2024ac3d} (T2V), CamCtrl~\citep{he2024cameractrl}, MotionCtrl~\citep{wang2024motionctrl}, and CAMI2V~\citep{zheng2024cami2v} (all I2V). These models represent different conditioned video generation models and are evaluated adhering to their default configurations. Following CAMI2V~\citep{zheng2024cami2v}, metrics include Camera Parameter Metrics (RotError, TransError, and CamMC) to quantify adherence to camera commands, and Fréchet Video Distance (FVD)~\citep{unterthiner2018towards} to assess perceptual realism. 

\noindent\textbf{Quantitative analysis.}
Our quantitative analysis on \nameGame\ reveals key insights and challenges. In the Text-to-Video task, AC3D showed basic camera control but high FVD, indicating the difficulty of generating high-fidelity, dynamic content with camera control and text prompts in complex scenes (\Cref{tab:benchmark_camctrl_eval}). For Image-to-Video models, CamCtrl achieves superior performance in both camera-controlled accuracy and video quality. However, all evaluated SOTAs still exhibit significant room for improvement across \nameGame, especially in simultaneously ensuring video generation quality and precise camera control. This highlights ongoing challenges and future research directions. 

\begin{figure*}[!t]
    \centering
    % \vspace{-1em}
    \includegraphics[width=\textwidth]{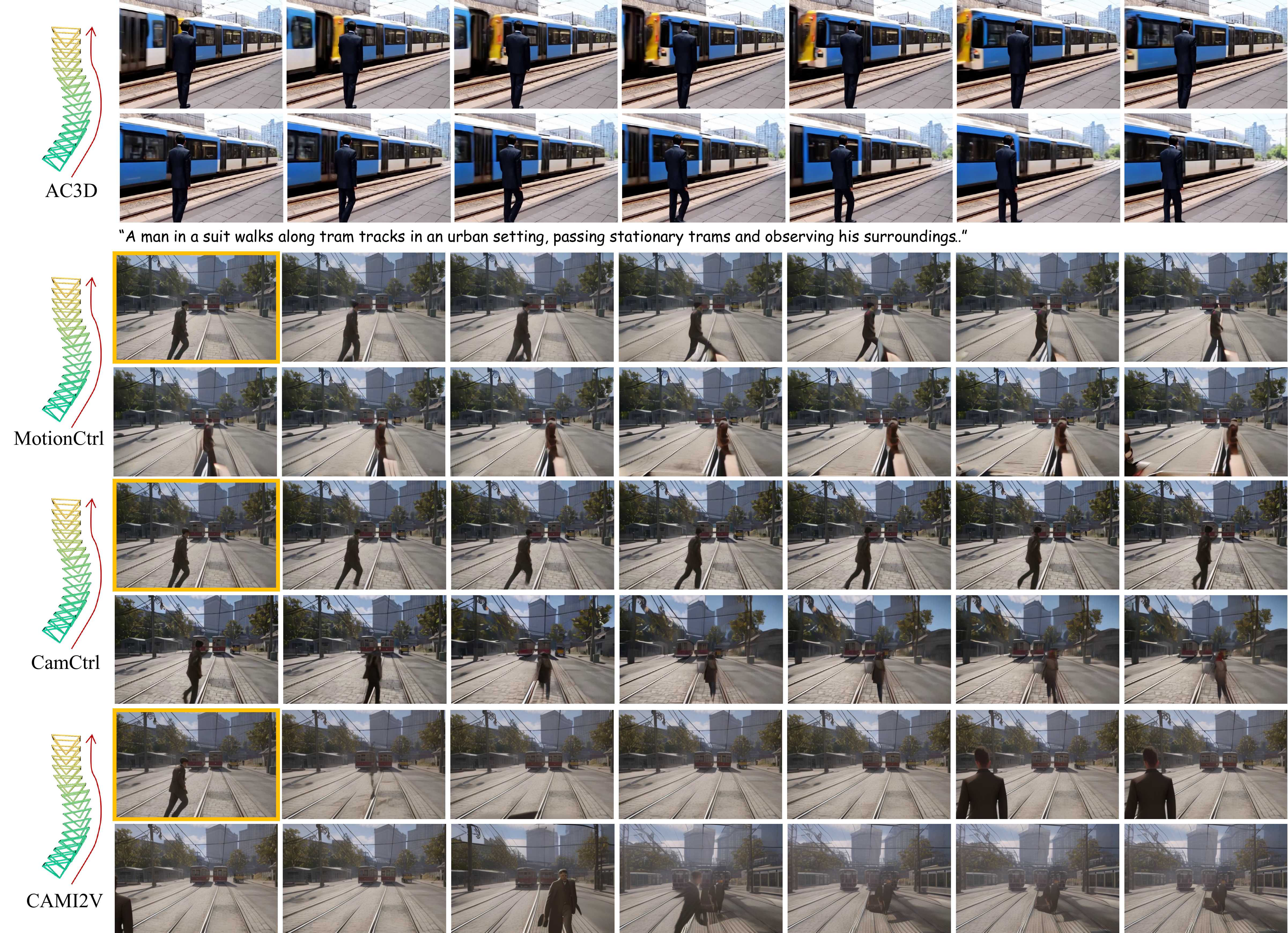}
    \captionsetup{justification=justified,singlelinecheck=false}
    \caption{~\textbf{Qualitative comparison of Camera-Controlled Video Generation} on \nameGame\ benchmark. In T2V setting, AC3D takes the text as a condition signal. In I2V setting, MotionCtrl, CamCtrl, CAMI2V takes the image as a condition signal. Condition images are the first images of each row.}
    \label{fig:benchmark_camctrl}
    % \vspace{-1em}
\end{figure*}

\noindent\textbf{Visual results.}
To visually demonstrate the challenges posed by the \nameGame\ benchmark, we present the qualitative results of various camera-controlled video generation models in \Cref{fig:benchmark_camctrl}. In the T2V setting, although AC3D~\citep{bahmani2024ac3d} generates semantically coherent video content, the depicted human motion is minimal, and the model fails to accurately follow the input camera trajectory. This highlights a fundamental limitation of current models in understanding and generating complex dynamic motions from abstract text instructions.
In the I2V setting, while the camera trajectory of CamCtrl's~\citep{he2024cameractrl} generated video aligns well with the input conditions, the visual quality of moving characters is blurry, and the overall video quality is poor. Similar quality degradation issues are observed in the outputs of MotionCtrl~\citep{wang2024motionctrl} and CAMI2V~\citep{zheng2024cami2v}. These results reveal the unique challenges of the \nameGame\ benchmark.

\section{Model Fine-tuning and Efficacy Validation}

Through comprehensive experiments, we systematically validate \name\ as a training source. We select baselines for two core tasks: 3D geometric foundation models and camera-controlled video generation models, and fine-tuned them using \name. The experimental results clearly demonstrate that models fine-tuned with \name\ consistently achieve significant performance improvements over their original published versions, powerfully confirming \name's capabilities in spatio-temporal modeling. 
\begin{table}[t]
\captionsetup{aboveskip=2pt}
    \centering
    % \vspace{-1em}
    \resizebox{1.0\columnwidth}!{
    \begin{tabular}{lcccccccc}
        \toprule[0.17em]
        \multirow{3}{*}{\textbf{Method}} &
        \multicolumn{2}{c}{\textbf{Sintel}} &
        \multicolumn{2}{c}{\textbf{Bonn}} &
        \multicolumn{2}{c}{\textbf{KITTI}} &
        \multicolumn{2}{c}{\textbf{NYU-v2}} \\
        \cmidrule(r){2-3} \cmidrule(r){4-5} \cmidrule(r){6-7} \cmidrule(r){8-9}
        &
        Abs Rel$\downarrow$ & $\delta<1.25\uparrow$ &
        Abs Rel$\downarrow$ & $\delta<1.25\uparrow$ &
        Abs Rel$\downarrow$ & $\delta<1.25\uparrow$ &
        Abs Rel$\downarrow$ & $\delta<1.25\uparrow$ \\
        \midrule%
        DUSt3R~\citep{dust3r_cvpr24} & 0.488 & 0.532 & 0.139 & 0.831 & 0.109 & 0.873 & \textbf{0.081} & \textbf{0.909} \\
        MonST3R~\citep{zhang2024monst3r} & \underline{0.402} & 0.525 & \underline{0.069} & \textbf{0.954} & \underline{0.098} & \underline{0.895} & 0.094 & 0.887 \\
        DUSt3R* & \textbf{0.370} & ~\underline{0.529} & \textbf{0.067} & \underline{0.948} & \textbf{0.088} & \textbf{0.932} & \underline{0.089} & \underline{0.902} \\
        \midrule
        CUT3R~\citep{wang2025continuous} & 0.420 & 0.520 & \textbf{0.058} & \textbf{0.967} & 0.097 & 0.914 & 0.081 & 0.914 \\
        CUT3R* & \textbf{0.408} & \textbf{0.522} & 0.075 & 0.944 & \textbf{0.087} & \textbf{0.935} & \textbf{0.075} & \textbf{0.920} \\
        \bottomrule[0.17em]
    \end{tabular}
    }
    \captionsetup{justification=justified,singlelinecheck=false}
    \caption{
        \textbf{Comparison of Original and Fine-tuned Models for Monocular Depth Estimation} on Sintel~\citep{Butler2012ANO}, Bonn~\citep{palazzolo2019arxiv}, KITTI~\citep{Geiger2013IJRR} and NYU-v2~\citep{silberman2012indoor}.  The notation * denotes models that have been fine-tuned on \name.}
    % \vspace{-1em}
    \label{tab:finetune_monodepth}
\end{table}
\begin{table}[t]
\captionsetup{aboveskip=2pt}
  \centering
%   % \vspace{-1em}
  \setlength{\tabcolsep}{3.2pt}
  \resizebox{\columnwidth}{!}{
  % \footnotesize
  \begin{tabular}{lccccccc}
    \toprule[0.17em]
    \multirow{3}{*}{\textbf{Method}} &
    \multirow{3}{*}{\textbf{Align}}  &
    \multicolumn{2}{c}{\textbf{Sintel}} &
    \multicolumn{2}{c}{\textbf{Bonn}}   &
    \multicolumn{2}{c}{\textbf{KITTI}}  \\

    \cmidrule(r){3-4}\cmidrule(r){5-6}\cmidrule(r){7-8}
    && Abs Rel $\downarrow$ & $\delta\!<\!1.25$ $\uparrow$
    & Abs Rel $\downarrow$ & $\delta\!<\!1.25$ $\uparrow$
    & Abs Rel $\downarrow$ & $\delta\!<\!1.25$ $\uparrow$ \\

    \midrule
    DUSt3R~\citep{dust3r_cvpr24} & \multirow{2}{*}{scale}
        & 0.652 & 0.436 & 0.151 & 0.839 & 0.143 & \textbf{0.814} \\
    DUSt3R* &
        & \textbf{0.512} & \textbf{0.456} & \textbf{0.083} & \textbf{0.920} & \textbf{0.135} & 0.800 \\
    
    \midrule
    CUT3R~\citep{wang2025continuous} & \multirow{2}{*}{scale}
        & 0.417 & 0.510 & 0.078 & 0.937 & 0.123 & 0.875 \\
    CUT3R* &
        & \textbf{0.396} & \textbf{0.516} & \textbf{0.078} & \textbf{0.938} & \textbf{0.107} & \textbf{0.907} \\

    \midrule
    DUSt3R~\citep{dust3r_cvpr24} &
        \multirow{2}{*}{\makecell[c]{scale\&shift}}
        & 0.570 & \textbf{0.493} & 0.152 & 0.835 & \textbf{0.135} & \textbf{0.818} \\
    DUSt3R* &
        & \textbf{0.520} & 0.480 & \textbf{0.084} & \textbf{0.914} & 0.136 & 0.808 \\
    
    \midrule
    CUT3R~\citep{wang2025continuous} & 
        \multirow{2}{*}{\makecell[c]{scale\&shift}}
        & 0.537 & 0.556 & 0.075 & 0.944 & 0.111 & 0.884 \\
    CUT3R* &
        & \textbf{0.314} & \textbf{0.574} & \textbf{0.067} & \textbf{0.964} & \textbf{0.103} & \textbf{0.912} \\
    \bottomrule[0.17em]
  \end{tabular}
  }% end \resizebox
  \captionsetup{justification=justified,singlelinecheck=false}
  \caption{\textbf{Comparison of Original and Fine-tuned Models for Video Depth Estimation} on Sintel~\citep{Butler2012ANO}, Bonn~\citep{palazzolo2019arxiv} and KITTI~\citep{Geiger2013IJRR}. The notation * denotes models that have been fine-tuned on \name.}
%   % \vspace{-1em}
  \label{tab:finetune_videodepth}
\end{table}

\subsection{Improving 3D Geometric Prediction with \name}

We select DUSt3R~\citep{dust3r_cvpr24}, CUT3R~\citep{wang2025continuous}, and Reloc3r~\citep{reloc3r} as our primary baselines and conduct fine-tuning experiments on subsets of \name.

The quantitative results confirm that models fine-tuned with \name\ consistently surpass their original performance across multiple critical tasks: monocular depth estimation (\Cref{tab:finetune_monodepth}), video depth estimation (\Cref{tab:finetune_videodepth}), and camera pose estimation. This outcome strongly demonstrates that \name's scale and diversity enable it to serve as a valuable large-scale training source, effectively enhancing the generalization capabilities and robustness of 3D geometric foundation models.

For monocular depth estimation (\Cref{tab:finetune_monodepth}), fine-tuned DUSt3R significantly outperformed its original baseline performance, even surpassing MonST3R, which is fine-tuned on multiple dynamic datasets~\citep{zheng2023point, wang2020tartanair, Mehl2023_Spring, sun2020scalability}. Similarly, CUT3R also showed improved performance after fine-tuning compared to the original baseline.

For video depth estimation (\Cref{tab:finetune_videodepth}), both DUSt3R and CUT3R exhibited enhanced performance after fine-tuning on \name, demonstrating \name's utility in improving temporal consistency.

For camera pose estimation, please refer to \textit{supplementary materials}.

\begin{table}[t]
\captionsetup{aboveskip=2pt}
\centering
% \vspace{-1em}
\resizebox{1.0\columnwidth}!{
\begin{tabular}{lcccccc}
\toprule[0.17em]
\multirow{2}{*}{\centering\textbf{Method}} &
\multirow{2}{*}{\centering\textbf{Benchmark}} &
\multirow{2}{*}{\centering\textbf{TransErr$\downarrow$}} &
\multirow{2}{*}{\centering\textbf{RotErr$\downarrow$}}  &
\multirow{2}{*}{\centering\textbf{CamMC$\downarrow$}}  &
\multicolumn{2}{c}{\textbf{FVD}} \\ 
\cmidrule(lr){6-7}
 &  &  &  & &
 {\footnotesize VideoGPT$\downarrow$} &
 {\footnotesize StyleGAN$\downarrow$} \\
\midrule
AC3D~\citep{bahmani2024ac3d}    & \multirow{2}{*}{RealEstate10K}   & 3.4433 & 0.6308 & 3.6615 & 479.320 & \textbf{409.795} \\
AC3D*   &  & \textbf{2.8648} & \textbf{0.5314} & \textbf{3.0518} & \textbf{472.683} & 416.948 \\
\midrule
AC3D~\citep{bahmani2024ac3d}    & \multirow{2}{*}{\nameGame}   & 6.2788 & 0.8867 & 6.6965 & 1745.778 & 1594.885 \\
AC3D*   &  & \textbf{4.1428} & \textbf{0.7610} & \textbf{4.4854} & \textbf{1437.247} & \textbf{1249.1858} \\

\bottomrule[0.17em]
\end{tabular}}
\captionsetup{justification=justified,singlelinecheck=false}
\caption{\small \textbf{Comparison of Original and Fine-tuned Models for Camera-Controlled Video Generation Evaluation} on RealEstate10K~\citep{zhou2018stereo} and \nameGame\ benchmark. The notation * denotes models that have been fine-tuned on \name.}
\label{tab:finetune_camctrl}
% \vspace{-1em}
\end{table}

\subsection{Enhancing Camera-Controlled Video Generation with \name}

Current public datasets for camera-controlled video generation models have significant limitations. For example, most datasets like RealEstate10K~\citep{zhou2018stereo} primarily consist of static scenes and relatively smooth camera movements, which hinders models' ability to generate dynamic video content.

To address this data bottleneck and validate \name's effectiveness, we select AC3D~\citep{bahmani2024ac3d} as our baseline and fine-tune it. Our experimental results further verify the finding from prior work (e.g., CAMERACTRL II~\citep{he2025cameractrl}), which highlight the critical importance of dynamic data for improving a model's camera-controlled capabilities.

The fine-tuned model is evaluated on two distinct benchmarks: a random subset of 150 video samples from the RealEstate10K test set and \nameGame\ benchmark, which consists of 200 video samples. For a fair comparison, all models are configured to output videos at a uniform resolution of 720 $\times$ 480 with a sequence length of 25 frames.

As shown in \Cref{tab:finetune_camctrl}, the model fine-tuned on \name\ significantly outperforms the original baseline model on both the RealEstate10K~\citep{zhou2018stereo} and \nameGame\ benchmarks. This outcome provides strong evidence that \name\ serves as an effective training resource, substantially enhancing the ability of controllable video generation models to follow precise camera-controlled instructions in complex and dynamic scenarios.
\section{Related Work}

\subsection{World Model Datasets}
The ability of models to perform world modeling is intrinsically linked to the availability of large-scale, high-quality spatio-temporal datasets. 

Static 3D datasets, such as ScanNet~\citep{dai2017scannet}, NYU-v2~\citep{silberman2012indoor}, and MegaDepth~\citep{MDLi18}, have advanced 3D reconstruction by providing precise geometric information. However, their static nature limits their utility for modeling motion and dynamic interactions. In video generation, large-scale video-text datasets~\citep{chen2024panda70m,Bain21,nan2024openvid,ju2024miradatalargescalevideodataset} offer rich semantic annotations but lack geometric information (e.g., depth, camera poses, optical flow), making them unsuitable for applications requiring precise 3D world modeling.

To bridge this gap, researchers have created dynamic real-world datasets like KITTI~\citep{Geiger2013IJRR} and Waymo~\citep{sun2020scalability} for autonomous driving, and Bonn~\citep{palazzolo2019arxiv}, HOI4D~\citep{Liu_2022_CVPR}, RH20T~\citep{fang2024rh20t}, and EPIC-Kitchens~\citep{Damen2018EPICKITCHENS} for human-robot interaction. While valuable, these datasets often suffer from a lack of scene diversity and noisy/sparse geometric annotations.

The sim-to-real gap has been significantly reduced due to the advancement of modern rendering technology~\citep{wang2020tartanair}.
Synthetic datasets have emerged as a valuable alternative, providing rich and precise ground-truth annotations. Pioneers like MPI Sintel~\citep{Butler2012ANO} are instrumental in optical flow research, but their small scale (e.g., an average sequence length of less than 50 frames) is insufficient for training large-scale foundation models. Other recent synthetic datasets, such as FlyingThings++~\citep{Mayer_2016_CVPR,harley2022particle}, TartanAir~\citep{wang2020tartanair}, Dynamic Replica~\citep{karaev2023dynamicstereo} and Spring~\citep{Mehl2023_Spring}, have made progress but still fall short in terms of scale, diversity, and modal richness compared to our self-collected \nameGame\ dataset, as shown in \Cref{tab:omniworld_comparison}.

The design of \name\ aims to systematically address these limitations. By integrating self-collected \nameGame\ dataset and several public datasets from various domains, we provide high-precision geometric annotations and rich spatio-temporal dynamics, enabling a more comprehensive evaluation and enhancement of world modeling.

\subsection{3D Geometric Foundation Models}
Recently, 3D geometric foundation models have emerged as a data-driven alternative to traditional methods like Structure-from-Motion (SfM), capable of directly predicting a scene's 3D structure in a single feed-forward pass. Early works like DUSt3R~\citep{dust3r_cvpr24} and MonST3R~\citep{zhang2024monst3r} operate on image pairs, requiring expensive global alignment for larger scenes. To overcome this, Fast3R~\citep{Yang_2025_Fast3R} enables simultaneous inference on thousands of images.

Other methods explore simplifying the learning task. FLARE~\citep{zhang2025flarefeedforwardgeometryappearance} decomposes the problem into separate pose and geometry prediction steps. CUT3R~\citep{wang2025continuous} is an online model that continuously updates its state from an image stream. VGGT~\citep{wang2025vggt} achieves superior performance through multi-task learning, while $\pi^3$~\citep{wang2025pi3} employs a permutation-equivariant architecture to remove the dependency on a fixed reference view. For monocular inputs, MoGe~\citep{wang2024moge,wang2025moge2} achieves accurate monocular geometry estimation by predicting affine-invariant point maps.

The performance of these methods is highly dependent on being trained on large-scale, multi-modal spatio-temporal datasets. When evaluated on the \nameGame\ benchmark, these methods show room for improvement, particularly when handling long sequences with highly dynamic, complex motions. By fine-tuning these models on \name, we achieve significant performance gains, powerfully demonstrating \name's value as an effective training resource for enhancing models' spatio-temporal modeling capabilities.

\subsection{Camera-Controlled Video Generation}
Camera-controlled video generation aims to empower users with the ability to control the camera within a generated video. Most methods in this field inject camera parameters (such as extrinsics or Plücker embeddings) into a pre-trained video diffusion model~\citep{blattmann2023stable,chen2023videocrafter1,yang2024cogvideox} with representative works including MotionCtrl~\citep{wang2024motionctrl}, CameraCtrl~\citep{he2024cameractrl}, CAMI2V~\citep{zheng2024cami2v}, and AC3D~\citep{bahmani2024ac3d}.

Despite this progress, these methods still struggle to generate dynamic content with complex camera control.
They are typically trained on datasets like RealEstate10K~\citep{zhou2018stereo} or DL3DV-10K~\citep{ling2024dl3dv}, which consist of static scenes with smooth camera motions. This data limitation inherently restricts a models' ability to handle dynamic scenes~\citep{he2025cameractrl}.

Our experiments confirm this limitation. When evaluated on \nameGame\ benchmark, which features rich dynamics and complex camera movements, these methods show considerable room for improvement in both visual quality and camera-controlled accuracy. By fine-tuning them on \name, their performance in dynamic scenes is significantly enhanced, demonstrating our dataset's value for improving models' spatio-temporal modeling capabilities.
\section{Conclusion}

In this work, we introduce \name, a large-scale, multi-domain, and multi-modal dataset designed to address the critical data bottleneck for world modeling. By integrating self-collected \nameGame\ dataset and several public datasets from various domains, we create a comprehensive data resource for world modeling. We demonstrate that \nameGame\ serves as a challenging benchmark for 3D geometric foundation models and camera-controlled video generation models, revealing the limitations of current SOTAs. Furthermore, we provide strong evidence that fine-tuning with \name\ significantly boosts the performance of these models, underscoring its value as a powerful training resource. We believe that \name\ will serve as a crucial data resource for the community, accelerating the development of more general and robust models for understanding and interacting with the real physical world.

\bibliography{main}

\appendix
% \clearpage
% \setcounter{page}{1}
\newpage
% \maketitlesupplementary

\appendix
% \section*{Appendix}
\addcontentsline{toc}{section}{Appendix}

\section{Overview}
\Cref{sec:suppl_data} discusses more details of \name. \Cref{sec:suppl_benchmark} and \Cref{sec:suppl_finetune} discuss more details of our benchmark and fine-tuning experiments.

\section{\name\ Dataset}
\label{sec:suppl_data}
\subsection{Data Statistics}

\begin{figure}[h]
    \centering
    \includegraphics[width=\linewidth]{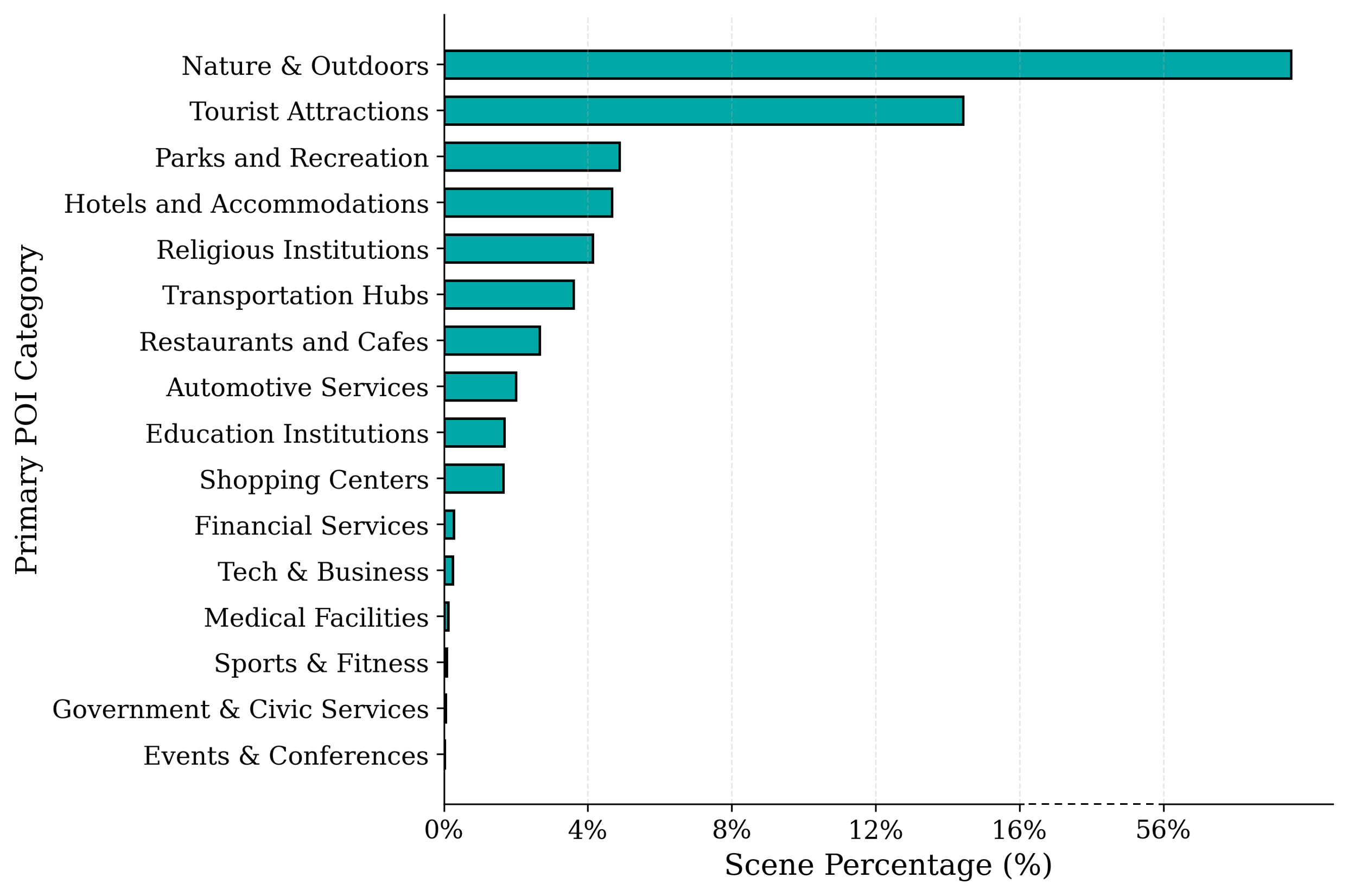}
    \captionsetup{justification=justified,singlelinecheck=false}
    \caption{~\textbf{The \nameGame\ distribution} of scene category (the primary POI locations).}
    \label{fig:statistic_poi_primary}
\end{figure}

To quantitatively analyze the scene diversity of \nameGame, we adopt the methodology from DL3DV~\citep{ling2024dl3dv} to classify and count scenes across 16 Point-of-Interest (POI) categories~\citep{ye2011exploiting}. The statistical results are shown in \Cref{fig:statistic_poi_primary}.
\nameGame\ encompasses a wide variety of scene categories, including "Nature \& Outdoors," "Tourist Attractions," "Parks and Recreation," and "Hotels and Accommodations." "Nature \& Outdoors" represents the largest share, reflecting its dominant presence in the dataset.
The distribution of these scene categories aligns with their prevalence in the real world and the characteristics of the games themselves. For instance, scenes related to "Government \& Civic Services" and "Events \& Conferences" are typically less frequent in games, leading to their lower representation in our dataset. These statistics further validate the richness and real-world attributes of \nameGame.

\begin{figure*}[!t]
    \centering
    % \vspace{-1em}
    \includegraphics[width=\textwidth]{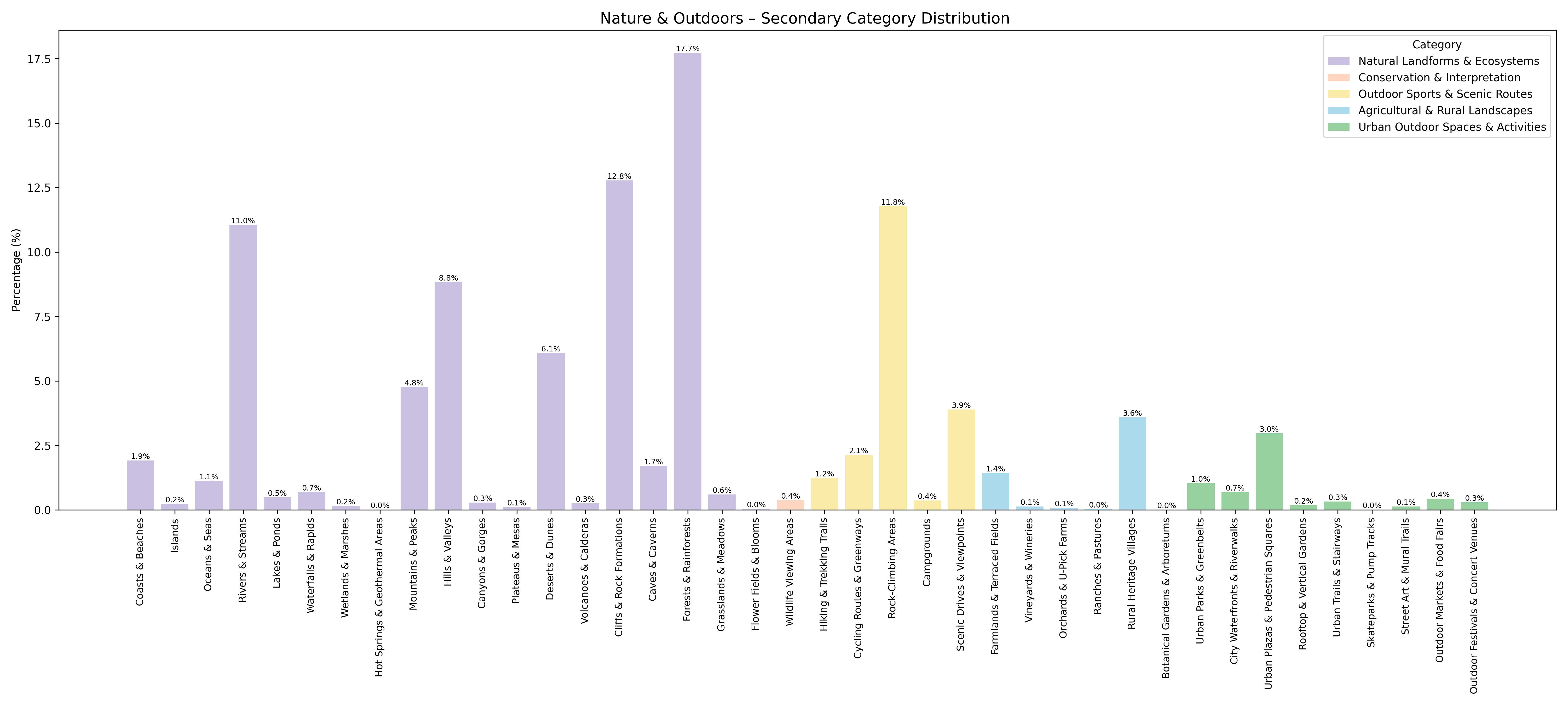}
    
    \captionsetup{justification=justified,singlelinecheck=false}
    \caption{~\textbf{Scene Diversity within the "Nature \& Outdoors" Category.} A quantitative breakdown of second- and third-level scene categories in \nameGame\ dataset, demonstrating the high internal diversity and distribution of natural environments.}
    \label{fig:statistic_outdoor_secondary_distribution}
    % \vspace{-1em}
\end{figure*}

To provide a more detailed analysis of the dominant "Nature \& Outdoors" scenes in \nameGame, we further subdivide this category into 5 second-level and 40 third-level categories. The detailed distribution is shown in \Cref{fig:statistic_outdoor_secondary_distribution}.
Our statistics reveal that "Natural Landforms \& Ecosystems" is the dominant second-level category. Within this category, scenes depicting "Forests \& Rainforests" and "Cliffs \& Rock Formations" are the most prevalent. "Outdoor Sports \& Scenic Routes" is the second-largest category, with scenes of "Rock-Climbing Areas" and "Scenic Drives \& Viewpoints" being particularly prominent.
Additionally, "Urban Outdoor Spaces \& Activities" and "Agricultural \& Rural Landscapes" also make up a small portion of the data.
These detailed statistics confirm that the "Nature \& Outdoors" scenes in \nameGame\ are not only abundant but also internally diverse. This rich composition provides a diverse data source for world modeling in complex natural environments.

\subsection{Ethics Statements}
To ensure compliance, we strictly adhere to the terms of use for relevant game content (e.g. Rockstar Games~\citep{rockstar_policy}), including usage for non-commercial purposes only and avoiding story spoilers. We also automatically remove UI elements and text information via a ReShade~\citep{reshade} plugin and manually filter specific scenes to ensure no unauthorized content is disclosed.

\section{\nameGame\ Benchmark}
\label{sec:suppl_benchmark}

\subsection{3D Geometric Prediction}

\noindent\textbf{Experiment Details.}
We adhere to the default configurations of each evaluated model. The entire evaluation process is conducted on a single A800 GPU.

For the monocular depth Estimation, we evaluate the first 200 frames of 18 test sequence from the \nameGame\ benchmark. Following the evaluation protocols of prior works~\citep{zhang2024monst3r, wang2025continuous, wang2025pi3}, we focus on scale-invariant monocular depth accuracy. The primary evaluation metrics are Absolute Relative Error (Abs Rel) and threshold accuracy ($\delta<1.25$). Under this setting, the depth map of each frame is independently aligned with its corresponding ground truth.

For the video depth estimation, we select the first 100 frames of the same test sequence from the \nameGame\ benchmark. To ensure a fair comparison across all models, we cap the input sequence length at 100 frames, as some models (e.g., FLARE~\citep{zhang2025flarefeedforwardgeometryappearance}) cannot handle longer sequences without errors. Similar to the mono depth estimation, we report Abs Rel and $\delta<1.25$. To more comprehensively evaluate depth consistency across video sequences, we provide results under two different alignment settings: (i) scale-only alignment (scale) and (ii) combined scale and translation alignment (scale \& shift). These settings test a model's depth estimation capabilities under different constraints, particularly in handling motion and viewpoint changes.

\begin{table}[t]
\captionsetup{aboveskip=2pt}
    \centering
    % \vspace{-1em}
    \resizebox{1.0\columnwidth}!{
    \begin{tabular}{lccccccccc}
        \toprule[0.17em]
        {\multirow{3}{*}{\textbf{Method}}} &
        \multicolumn{3}{c}{\textbf{Sintel}} &
        \multicolumn{3}{c}{\textbf{TUM-dynamics}} &
        \multicolumn{3}{c}{\textbf{ScanNet}} \\
        \cmidrule(r){2-4} \cmidrule(r){5-7} \cmidrule(r){8-10}
        &
        ATE$\downarrow$ & RPE trans$\downarrow$ & RPE rot$\downarrow$ &
        ATE$\downarrow$ & RPE trans$\downarrow$ & RPE rot$\downarrow$ &
        ATE$\downarrow$ & RPE trans$\downarrow$ & RPE rot$\downarrow$ \\
        \midrule
        CUT3R~~\citep{wang2025continuous} & 0.210 & 0.071 & \textbf{0.627} & 0.045 & 0.014 & 0.441 & 0.096 & 0.022 & 0.733 \\
        CUT3R* & \textbf{0.178} & \textbf{0.055} & 0.651 & \textbf{0.041} & \textbf{0.013} & \textbf{0.374} & \textbf{0.095} & \textbf{0.022} & \textbf{0.604} \\
        \bottomrule[0.17em]
    \end{tabular}
    }
    \captionsetup{justification=justified,singlelinecheck=false}
    \caption{
        \textbf{Comparison of Original and Fine-tuned Models for Camera Pose Estimation} on Sintel~~\citep{Butler2012ANO}, TUM-dynamics~~\citep{sturm2012benchmark} and ScanNet~~\citep{dai2017scannet}. The notation * denotes models that have been fine-tuned on \name.
    }
    % \vspace{-1em}
    \label{tab:finetune_relpose_cut3r}
\end{table}
\begin{table}[t]
\captionsetup{aboveskip=2pt}
    \centering
    % \vspace{-1em}
    \resizebox{1.0\columnwidth}!{
    \begin{tabular}{lcccccc}
        \toprule[0.17em]
        {\multirow{3}{*}{\textbf{Method}}} &
        \multicolumn{3}{c}{\textbf{DynPose-100K}} &
        \multicolumn{3}{c}{\textbf{OmniWorld-CityWalk}} \\
        \cmidrule(r){2-4} \cmidrule(r){5-7}
        &
        AUC@5$\uparrow$ & AUC@10$\uparrow$ & AUC@20$\uparrow$ &
        AUC@5$\uparrow$ & AUC@10$\uparrow$ & AUC@20$\uparrow$ \\
        \midrule
        Reloc3r~\citep{reloc3r} & 6.9 & 15.4 & 27.1 & 33.3 & 49.4 & 63.1 \\
        Reloc3r* & \textbf{14.4} & \textbf{25.5} & \textbf{37.8} & \textbf{42.5} & \textbf{58.0} & \textbf{70.3} \\
        \bottomrule[0.17em]
    \end{tabular}
    }
    \captionsetup{justification=justified,singlelinecheck=false}
    \caption{
        \textbf{Comparison of Original and Fine-tuned Models for Relative Camera Pose Evaluation} on DynPose-100K~\citep{rockwell2025dynamic}, OmniWorld-CityWalk\citep{li2025sekai}. The notation * denotes models that have been fine-tuned on \name.
    }
    % \vspace{-1em}
    \label{tab:finetune_relative_camera_pose}
\end{table}

It is important to note that since the benchmark data is included in the training set of $\pi^3$~\citep{wang2025pi3}, we did not evaluate it in our benchmark.

\subsection{Camera-Controlled Video Generation}
\noindent\textbf{Experiment Details.}
AC3D~\citep{bahmani2024ac3d} uses CogVideoX-5B~\citep{yang2024cogvideox} as base T2V model, it generates 25 frames per inference at a resolution of 480 $\times$ 720.
CamCtrl~\citep{he2024cameractrl} and MotionCtrl~\citep{wang2024motionctrl} use Stable Video Diffusion (SVD)~\citep{blattmann2023stable} as base I2V model and generate 14-frame video sequences at a resolution of 320 $\times$ 512.
CAMI2V~\citep{zheng2024cami2v} uses DynamiCrafter~\citep{xing2023dynamicrafter} as base I2V model. It generates 16-frame video sequences at a resolution of 320 $\times$ 512. For a fair comparison with CamCtrl and MotionCtrl, we use the first 14 frames of its generated videos for evaluation.
We use $\pi^3$~\citep{wang2025pi3} to get camera poses of the generated videos.
All methods are evaluated on an A800 GPU.

\section{Model Fine-tuning}
\label{sec:suppl_finetune}
\subsection{Camera Pose Estimation.}
Following~\citep{wang2025continuous, wang2025pi3}, we report the Absolute Trajectory Error (ATE), Relative Pose Error for translation (RPE trans), and Relative Pose Error for rotation (RPE rot) on Sintel~\citep{Butler2012ANO}, TUM-dynamics~\citep{sturm2012benchmark} and ScanNet~\citep{dai2017scannet}.
The results in \Cref{tab:finetune_relpose_cut3r} show that CUT3R's performance notably improved after fine-tuning on \name\ in camera pose estimation.

Following~\citep{reloc3r}, we assess performance with three indicators: AUC@5/10/20, which measure the area under the pose accuracy curve. This curve is based on minimum thresholds of 5, 10, and 20 degrees for rotation and translation angular errors. Reloc3r demonstrated substantial improvements in its ability to estimate dynamic camera poses after fine-tuning on \name\ in relative camera pose evaluation (\Cref{tab:finetune_relative_camera_pose}).

\subsection{Implementation Details}
We conduct comprehensive fine-tuning experiments on several SOTAs to validate the efficacy of our \name\ as a training resource. All experiments are performed on 8 NVIDIA A800 GPUs.

\noindent\textbf{DUSt3R~\citep{dust3r_cvpr24}.}
For fine-tuning, we use \nameGame\ alongside a portion of DUSt3R's original training sets, including ARKitScenes~\citep{dehghan2021arkitscenes}, MegaDepth~\citep{MDLi18}, and Waymo~\citep{sun2020scalability}. We load the pre-trained weights of DUSt3R and performed full fine-tuning. The model is fine-tuned on images with random resolutions (e.g., 288×512, 384×512, 336×512). The training runs for 40 epochs, with each epoch consisting of 800 iterations. We use the AdamW optimizer with an initial learning rate of $2.5 \times 10^{-5}$ and a weight decay of 0.05. Each GPU had a batch size of 7, with each batch containing two images.

\noindent\textbf{CUT3R~\citep{wang2025continuous}.}
We fine-tune CUT3R using \nameGame\ and a subset of its original training data, including CO3Dv2~\citep{reizenstein2021common}, WildRGBD~\citep{xia2024rgbd}, ARKitScenes~\citep{dehghan2021arkitscenes}, Waymo~\citep{sun2020scalability}, and TartanAir~\citep{wang2020tartanair}. We load the pre-trained weights and follow the training strategy from CUT3R's training stage 3. We fine-tune on higher-resolution images with varied aspect ratios, setting the maximum side to 512 pixels. The encoder is frozen, with only the decoder and heads being trained on longer sequences of 4 to 64 views. The model is fine-tuned for 2,000 iterations with a total batch size of 96 and a learning rate of $1.0 \times 10^{-6}$, optimized by AdamW with a weight decay of 0.05.

\noindent\textbf{Reloc3r~\citep{reloc3r}.}
For fine-tuning Reloc3r, we utilize \nameGame, \textit{OmniWorld-CityWalk}, \textit{OmniWorld-HoloAssist}, and \textit{OmniWorld-EpicKitchens}, along with a portion of its original training sets, including CO3Dv2~\citep{reizenstein2021common}, ARKitScenes~\citep{dehghan2021arkitscenes}, Scannet++~\citep{yeshwanth2023scannet++}, BlendedMVS~\citep{yao2020blendedmvs}, and MegaDepth~\citep{MDLi18}. We load the pre-trained weights, freeze the ViT encoder, and only update the weights for the decoder and pose regression head. Fine-tuning is performed on images of random resolutions, including 288 $\times$ 512, 384 $\times$ 512, and 336 $\times$ 512. The model is trained for 80 epochs, with each epoch comprising 400 iterations. We use the AdamW optimizer with a learning rate of $5.0 \times 10^{-6}$ and a weight decay of 0.05. Each GPU has a batch size of 32, with each batch containing two images.

\noindent\textbf{AC3D~\citep{bahmani2024ac3d}.}
We fine-tune AC3D using \nameGame, \textit{OmniWorld-EpicKitchens}, \textit{OmniWorld-HOI4D}, \textit{OmniWorld-HoloAssist}, \textit{OmniWorld-EgoExo4D}, and \textit{OmniWorld-EgoDex}, as well as the original training set, RealEstate10K~\citep{zhou2018stereo}. We load the pre-trained weights of the AC3D ControlNet~\citep{zhang2023adding}, which is based on CogVideoX-5B~\citep{yang2024cogvideox}. Only the ControlNet model is fine-tuned, with other network structures frozen. The fine-tuning is performed on video clips of 49 frames with a resolution of 352 $\times$ 640. The model is fine-tuned for 6,000 iterations with a total batch size of 8 and a learning rate of $5.0 \times 10^{-5}$, optimized by AdamW with a weight decay of 0.0001.

\begin{figure*}[!t]
    \centering
    % \vspace{-1em}
    \includegraphics[width=\textwidth]{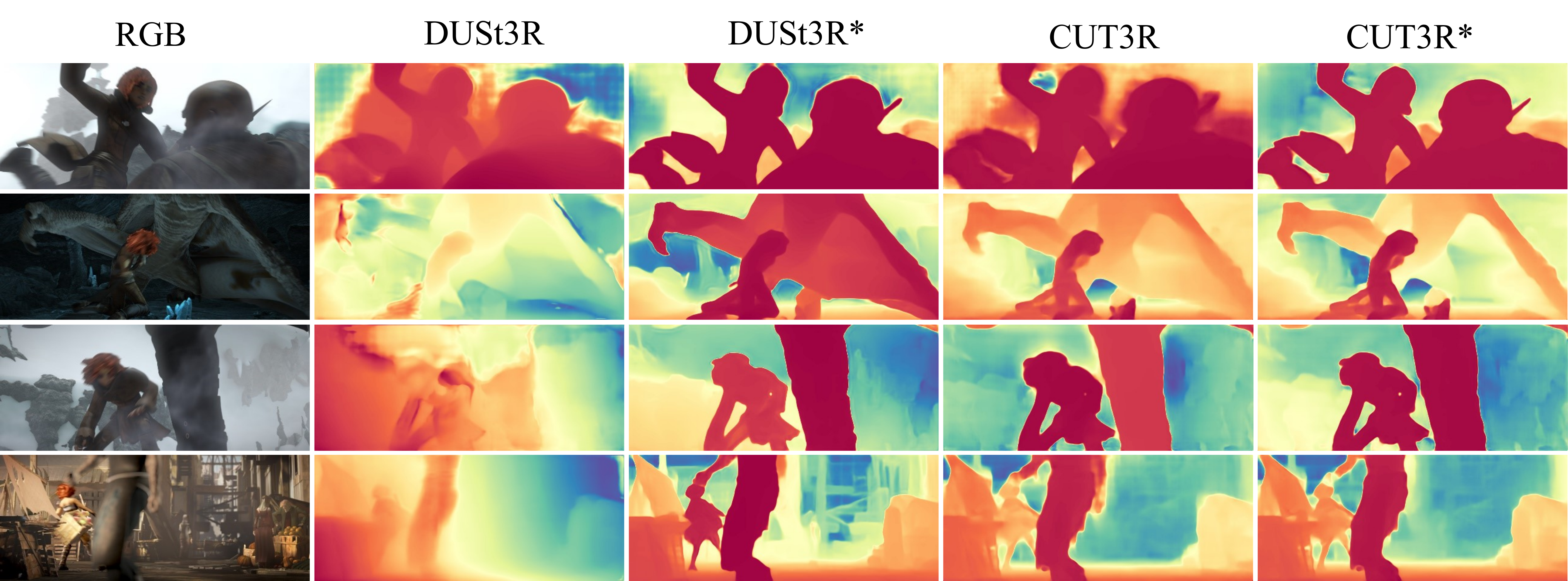}
    \captionsetup{justification=justified,singlelinecheck=false}
    \caption{~\textbf{Qualitative comparison of Original and Fine-tuned Models for Video Depth Estimation} on the Sintel~\citep{Butler2012ANO}. * denotes models that have been fine-tuned on \name. After fine-tuning, both models recover finer geometric details and produce more accurate depth maps, highlighting the efficacy of \name\ as a geometric supervision source.}
    \label{fig:finetune_videodepth_vs}
    % \vspace{-1em}
\end{figure*}

\begin{figure*}[!t]
    \centering
    % \vspace{-1em}
    \includegraphics[width=\textwidth]{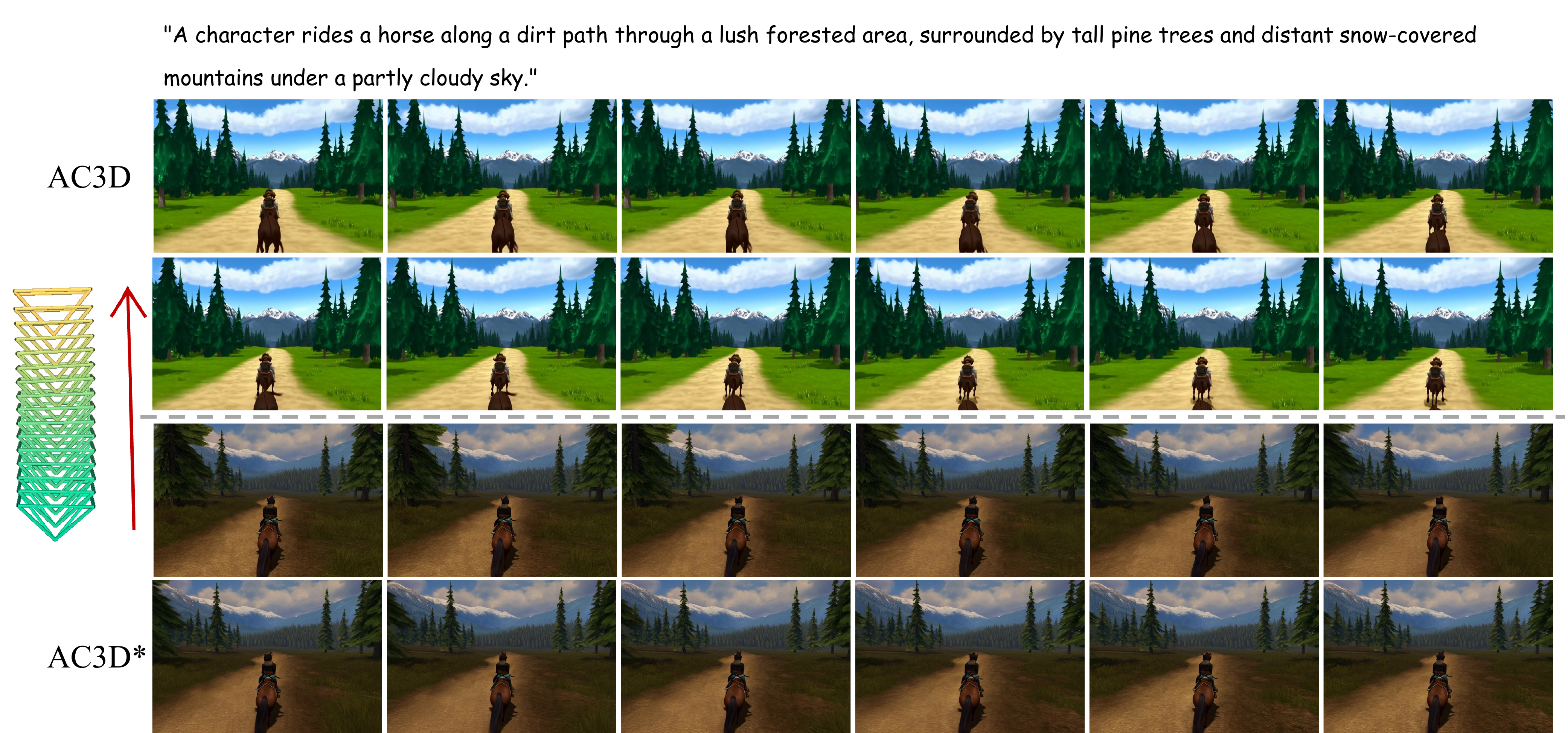}
    \captionsetup{justification=justified,singlelinecheck=false}
    \caption{~\textbf{Qualitative comparison of Original and Fine-tuned Models for Camera-Controlled Video Generation.} * denotes models that have been fine-tuned on \name. The visualizations show that fine-tuning with our dataset significantly improves the model's ability to generate videos that more accurately follow camera trajectories and maintain higher temporal consistency for moving objects.}
    \label{fig:finetune_camctrl_vs}
    % \vspace{-1em}
\end{figure*}

\subsection{Visual Results.}

\Cref{fig:finetune_videodepth_vs} provides a qualitative comparison of DUSt3R~\citep{dust3r_cvpr24} and CUT3R~\citep{wang2025continuous} on the Sintel~\citep{Butler2012ANO} subset of the Video Depth Estimation benchmark, evaluated both before and after fine-tuning on \name. After fine-tuning, both models recover finer geometric details and generate more accurate depth maps. These results indicate that \name\ offers strong geometric supervision and can substantially enhance a model's geometric prediction capability.

\Cref{fig:finetune_camctrl_vs} presents a visual comparison of AC3D~\citep{bahmani2024ac3d} on the \nameGame\ benchmark before and after fine-tuning on the \name\ dataset for the camera-controlled video generation task. The visualizations clearly show that after fine-tuning, the generated videos more closely follow the desired camera trajectory and exhibit higher temporal consistency for moving objects. This demonstrates that \name\ can significantly enhance a model's ability to model dynamics.

\end{document}